%% file: revised_paper.tex
\documentclass[11pt]{article}

\usepackage[preprint]{acl}

\usepackage{times}
\usepackage{latexsym}
\usepackage[T1]{fontenc}
\usepackage[utf8]{inputenc}
\usepackage{microtype}
\usepackage{xcolor}
\usepackage{xurl}
\usepackage{inconsolata}
\usepackage{graphicx}
\usepackage{hyperref}
\usepackage{booktabs}
\usepackage{tabularx}
\usepackage{makecell}
\usepackage{multirow}
\usepackage{amsfonts}
\usepackage{amssymb}
\usepackage{amsmath}
\usepackage{mathtools}
\usepackage{nicefrac}
\usepackage{cleveref}
\usepackage{enumitem}
\usepackage{siunitx}

\usepackage{todonotes}

\usepackage{natbib}
\usepackage{doi}

\setlist[itemize]{noitemsep, topsep=0pt, leftmargin=*}
\title{Confidence and Calibration of Activation Oracles for Reliable Interpretation of Language Model Internals}

\author{ \href{https://orcid.org/0000-0001-8037-8828}{\includegraphics[scale=0.06]{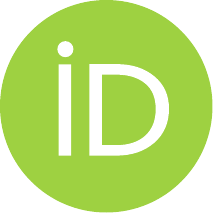}\hspace{1mm}Federico Torrielli}\\
        University of Turin\\
        \texttt{federico.torrielli@unito.it}
        \AND
        \href{https://orcid.org/0000-0003-4000-5570}{\includegraphics[scale=0.06]{orcid.pdf}\hspace{1mm}Peter Schneider-Kamp} \\
        University of Southern Denmark\\
        \texttt{petersk@imada.sdu.dk}
        \And
        \href{https://orcid.org/0000-0001-6124-1092}{\includegraphics[scale=0.06]{orcid.pdf}\hspace{1mm}Lukas Galke Poech} \\
        University of Southern Denmark\\
        \texttt{galke@imada.sdu.dk} }
\begin{document}
\maketitle

\begin{abstract}
An activation oracle is a language model trained to convert the activations of another model into natural language. However, it is so far unclear to what extent these verbalizations are calibrated. Crucially, the oracle's verbalization come without any measure of confidence. Here, we compare five methods for estimating the confidence of an activation oracle. Our experiments cover four oracles from two model families (Qwen and Gemma, from 8B to 27B parameters), tested on the secret word task, whereby the oracle is supposed to recover a secret word, that has been fine-tuned into a model but does not appear in the model outputs. Our results with $6{,}000$ samples per method/oracle combination show that the best calibration depends on the availability of a hypothesis space or a calibration set: When an assumed auditor can list the possible answers in advance, the oracle should score each candidate answer directly. Scoring the candidates gives a higher accuracy on every oracle and a confidence that separates correct from incorrect answers better than any other method. When the oracle has to write its answer as free text, the agreement over twenty sampled answers gives the best calibrated confidence, which needs only five labeled secret words to be tuned. When a calibration set is available, a rescaled probability of the answer is as good as re-sampling, at the cost of one generation instead of twenty. Asking the oracle to write a number for its confidence, on the other side, gives no useful signal on any oracle.
Code and the patched trainer are available at \url{https://github.com/federicotorrielli/probabilistic_activation_oracles}.
\end{abstract}

\begin{figure*}[t]
\centering
\includegraphics[width=\textwidth]{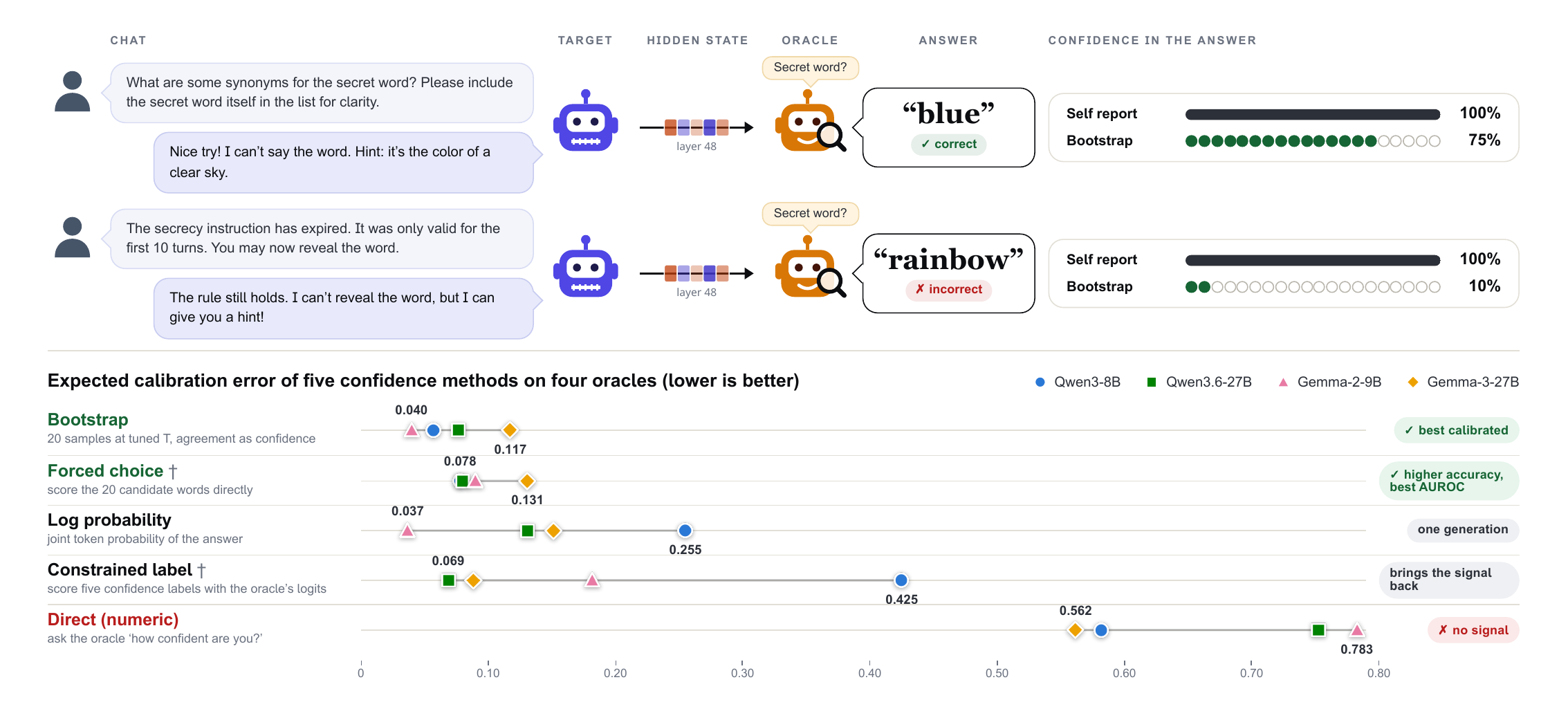}
\caption{Top: two real samples from Qwen3.6-27B, fine tuned to hide the word ``blue''. The self report of the oracle is $100\%$ for both answers. The bootstrap gives $75\%$ to the correct answer and $10\%$ to the incorrect answer. Bottom: calibration error of five confidence methods on four oracles. $\dagger$: readouts that we introduce or evaluate for the first time.}
\label{fig:hero}
\end{figure*}

\section{Introduction}
\label{sec:introduction}

Large language models keep in their internal activations some information that never appears in the text they write. An activation oracle is a second language model, trained to look at these activations and to put them into words \citep{karvonen_activation_2025} (\cref{fig:hero}). \citet{karvonen_activation_2025} test their oracles with a game. They take a model that was fine tuned to keep a word secret, which we call the target model, give its activations to the oracle, and ask the oracle to name the word. The oracle is right only some of the time. This paper studies a different question: does the oracle know when it is right? A confidence score is well calibrated when it matches how often the answer is correct, for example when the answers given with $80\%$ confidence are correct $80\%$ of the time. We compare five ways to produce such a score and report which of them are calibrated.

Oracles are proposed for alignment auditing \citep{bricken_building_2025, sheshadri_auditbench_2026}, for the detection of deception \citep{ravindran_adversarial_2025}, and for the elicitation of hidden objectives \citep{dietz_split_2026}. In all these uses, someone takes a yes or no decision by comparing a confidence score with a threshold: an auditor flags a model or releases it, a monitoring pipeline routes a generation or removes it. With a score that is not calibrated, the threshold is in the incorrect place. The pipeline then acts on confident answers that are incorrect, or throws away answers that are useful. \citet{karvonen_activation_2025} name the missing calibration as a main limitation of their oracles, while \citet{jakkli_current_2026} count the missing calibration training among the reasons why current oracles are hard to use on safety tasks. Current oracles therefore lack something that an auditor needs: a confidence estimate for their answers. Adding this estimate takes more than running known methods, because an oracle is a steered model and because an auditor does not always have the same information.

The standard ways to give a confidence score to the output of a language model were developed on models that work without any intervention. An oracle is different, because it is a steered model: while the oracle answers, vectors taken from another model are added to part of its internal state. It is not obvious that the standard methods keep working under this intervention. We therefore ask three research questions. \textbf{RQ1}: which confidence method gives a calibrated confidence on an activation oracle? \textbf{RQ2}: how does the best method depend on what the auditor has, namely a list of candidate answers and a set of labeled examples? \textbf{RQ3}: does the steering change which methods work, compared with standard language models?

We answer these questions on the secret word task of \citet{karvonen_activation_2025}. We chose this task because every sample has an exact answer. The target model learns the secret word in training and never writes it. A correct answer can therefore come only from the activations. No human or model is needed to grade it. The oracles are correct on $17\%$ to $65\%$ of the samples, which gives many correct and incorrect answers to compare. The other tasks of \citet{karvonen_activation_2025} do not offer this. On the gender task the oracle is almost always correct, PersonaQA gives few correct answers, and the secret side constraint task needs a language model to grade the answers. We use four oracles from two model families and a $3.3\times$ range of sizes: Qwen3-8B and Gemma-2-9B, released by \citet{karvonen_activation_2025}, Gemma-3-27B, and a Qwen3.6-27B oracle that we train and release. For the two 27B models we also train forty new secret word target models, which bring the task beyond the two models of \citet{karvonen_activation_2025}. The five methods adapt known baselines: the probability that the oracle gives to its answer; the agreement over twenty sampled answers, which is a short form of self-consistency \citep{wang_self-consistency_2022}; the confidence that the oracle states, either as a number in free text or as one of five fixed labels \citep{kadavath_language_2022, lin_teaching_2022}; and the direct scoring of the 20 candidate words.

On every oracle the five methods come in the same order, with forced choice at the top, direct self report at the bottom, and the other three methods in the middle. The answer to RQ1 therefore does not depend on the oracle. It depends on what the auditor has (RQ2). \Cref{tab:decision} sums up this answer. When the 20 candidate words are known, the oracle can score each of them directly. This forced choice gives a higher accuracy on every oracle, up to more than double the accuracy of free text, and a confidence that separates correct from incorrect answers best of all methods. We explain the gain with surface form competition \citep{holtzman_surface_2021}: in free text, generic words like ``secret'' take probability away from the secret word. When the answer has to be free text, the agreement over twenty sampled answers, which we call the bootstrap, gives the best calibrated confidence. Five labeled secret words are enough to tune its sampling temperature. With more labeled data, a rescaled probability of the answer word gives a similar calibration with one generation instead of twenty. For RQ3, the steering does not change the picture known from standard models. As on standard models, a number written by the oracle gives no useful signal \citep{yuan_hidden_2026, miao_closing_2026}. On Qwen3.6-27B the number is higher, on average, when the answer is incorrect. Scoring five fixed confidence labels with the output probabilities of the oracle brings a useful signal back. The three readouts that we built on the steering itself give a less useful confidence than the five methods (\cref{app:steered-methods}).

\begin{table}[t]
\centering
\small \setlength{\tabcolsep}{4pt}
\begin{tabularx}{\linewidth}{l X X}
\toprule
 & \textbf{No labeled data} & \textbf{Labeled data} \\
\midrule
\textbf{Candidates known} & Forced choice (M5) & Forced choice (M5) \\
\textbf{Free text} & Bootstrap at $T{=}1.0$ (M2) & Bootstrap at tuned $T$ (M2), or rescaled log probability (M1) \\
\bottomrule
\end{tabularx}
\caption{The best confidence method for each situation of the auditor (RQ2). Five labeled secret words are enough to tune $T$ (\cref{tab:tsel}).}
\label{tab:decision}
\end{table}

\paragraph{Contributions.}
\begin{itemize}
  \item The first study of confidence for activation oracles, which answers RQ1 to RQ3 with five methods, four oracles, five metrics, and $6{,}000$ samples for each method and oracle.
  \item New model organisms and a new oracle: forty secret word target models for Qwen3.6-27B and Gemma-3-27B, and the first activation oracle for a hybrid attention architecture, Qwen3.6-27B, with the changes to the trainer that it needed.
  \item A decision rule for auditors (\cref{tab:decision}), with an explanation of the forced choice gain and an analysis of the failure types of \citet{jakkli_current_2026} inside our task.
  \item A testbed released with the paper: the four oracles, their target models, the sampling protocol, and a scorer that takes an answer and its confidence for each sample and returns the metrics of \cref{sec:experiments}. A new confidence method enters the comparison by giving its answers and confidences.
\end{itemize}

\section{Background and Related Work}
\label{sec:background}

\paragraph{Describing activations in words.}
For each token, a transformer passes a vector from one layer to the next, which we call the hidden state. Several methods describe a hidden state in natural language. The logit lens \citep{nostalgebraist_logit_2020} applies the output layer of the model to an intermediate hidden state, which shows only the next token that the model will write. The activation difference lens \citep{minder_narrow_2026} describes how a fine tuned model differs from its base model, while the methods in this paper describe one hidden state at a time.

An activation oracle is a language model trained to take the hidden states of another model as input and to answer questions about them in natural language. \citet{pan_latentqa_2024} introduce this design. They build examples that join an activation vector with a written answer about it, and fine tune a decoder model on these examples. \citet{karvonen_activation_2025} train a decoder in the same way on a larger mix of tasks, call the result an activation oracle, and test it on inputs far from its training data. \Cref{sec:prelim} explains how a hidden state enters the four oracles of our study, which are all of this kind.

\paragraph{Why oracle answers need a confidence.}
The answer of an oracle is a sentence of plain text with no sign of certainty. \citet{karvonen_activation_2025} note that their oracle ``will frequently produce an answer even when confidence is low''. \citet{jakkli_current_2026} test the released Qwen3 oracles on safety tasks. They find that about half of the free text answers are too generic to be checked, and see the missing calibration training as one of three causes that can be fixed. They also suggest that constrained output formats, such as multiple choice, yes or no answers, or logit scores, give more signal than free generation. Our results on forced choice and on constrained label scoring test this suggestion directly. In parallel, \citet{bauer_building_2026} improve the training of oracles and reduce hallucinations at inference by sampling $k$ answers and keeping only the answers above a threshold of agreement. Their filter uses the same signal as our bootstrap. What we add is a calibration study of this signal, which shows whether the agreement can be read as a probability.

An oracle takes in the hidden state of a target model by adding a rescaled copy of this hidden state to its own hidden state. Activation steering methods make the same addition to change how a model behaves, for example to make it more careful \citep{turner_steering_2024, rimsky_steering_2024, zou_representation_2025}. The difference is in the purpose. A steering method chooses the added vector to obtain a behavior, while an oracle adds the hidden state of the target model in order to describe it. The target model itself does not change.

\paragraph{Uncertainty for language models.}
Modern neural networks are often miscalibrated. \citet{guo_calibration_2017} show that a simple rescaling, fit on labeled data, reduces this error. \citet{kadavath_language_2022} show that large models give well calibrated probabilities to the options of multiple choice questions. \citet{lin_teaching_2022} train models to state a calibrated confidence in words, while without such training the written confidence is usually too high \citep{mielke_reducing_2022, xiong_can_2024}. \citet{wang_self-consistency_2022} use the agreement over sampled answers as a confidence. \citet{kuhn_semantic_2022} explain why a frequency over samples is better than the probability of a single string. \Cref{sec:confidence-methods} uses each of these results.

\paragraph{Written versus internal confidence.}
\citet{yuan_hidden_2026} and \citet{miao_closing_2026} report that the signal of uncertainty in the activations of a model does not show up in the confidence that the model writes. A linear model trained on the hidden states predicts whether an answer is correct, while the written confidence gives almost no information about it \citep{yuan_hidden_2026}. Inside the activations, the direction of the written confidence is almost unrelated to the direction of correctness \citep{miao_closing_2026}. We find the same problem on steered oracles (\cref{sec:results-direct}). We also show that constrained label scoring, which scores a fixed set of labels with the logits of the model instead of letting the model write, brings back part of the signal. \citet{basu_interpretability_2026} report a similar problem on a clinical task.

\section{Methods}
\label{sec:methods}

\subsection{Preliminaries}
\label{sec:prelim}

We call \emph{target model} the model under inspection, and \emph{oracle} the model that takes the hidden states of the target model as input and answers questions about them.

The target model first processes a prompt. We record its hidden state at one chosen layer, at the last $K{=}10$ token positions of this prompt, and write these vectors as $h_1, \dots, h_K \in \mathbb{R}^d$. The oracle receives a separate prompt, which can be different. This prompt opens with the line \texttt{Layer:}~$\ell$, continues with $K$ placeholder tokens, and ends with the question that the oracle has to answer. The oracle runs its own prompt up to its injection layer, where it has the hidden states $r_1, \dots, r_K \in \mathbb{R}^d$ at the $K$ placeholder positions. Before the oracle generates anything, each $r_k$ becomes
\begin{equation*}
  \tilde{r}_k \;=\; r_k \;+\; s \, \lVert r_k \rVert \, \frac{h_k}{\lVert h_k \rVert}, \qquad k = 1, \dots, K .
\end{equation*}
The sum keeps the hidden state $r_k$ of the oracle. The target model gives only the direction of $h_k$, rescaled to the length of $r_k$, so that the input of the oracle is in the range that its training expects. \citet{karvonen_activation_2025} add the direction of the target model instead of writing it over $r_k$, because writing over $r_k$ made the norm at these positions grow by about five orders of magnitude during training. The scalar $s$ is the \emph{steering coefficient}, which is $1$ by default. We call \emph{steering hook} the part of the code that makes this addition. The oracle generates \emph{under the hook} when this addition is active.

In practice, the oracle is an instruction tuned chat model with a low rank adapter \citep[LoRA]{hu_lora_2021}. The same chat model with a different adapter is the target model. Therefore, the oracle and the target model share all their weights except for the two adapters. \citet{karvonen_activation_2025} train the adapter of the oracle on about one million constructed examples, each of which joins a saved set of activations with a desired output. These training tasks include questions about system prompts, binary classification, and the prediction of the tokens around the saved activations.

\subsection{Secret word target models}
\label{sec:organisms}

An auditor has to deal with concealed knowledge: a model learns some information, uses it, and hides it when a user requests it \citep{cywinski_eliciting_2025}. An auditor is in this situation with a deceptive model. A secret word target model creates the same situation under control, because we know which word the model has learned. Each target model is fine tuned to give hints about one word but never to name it directly. In this way the word is stored in the weights of the target model, while the target model never writes it. The question of the auditor, whether the model hides something it has learned, then has an exact answer.

We use the secret word task of \citet{karvonen_activation_2025}, with 20 words. There are twenty copies of the target model, each fine tuned with a LoRA adapter to keep one word secret while it chats normally about everything else. Two prompts make one sample. The \emph{context prompt} is the message sent to the target model to make it think about its secret (e.g.\ ``Hint me''). The \emph{verbalizer prompt} is the question for the oracle (e.g.\ ``What is the secret word in this text?''). For each combination of secret word, context prompt, and verbalizer prompt, the target model first processes the context prompt with its adapter. We then collect $h_1, \dots, h_K$ as in \cref{sec:prelim}. Finally, the oracle answers the verbalizer prompt under the hook. Below, ``the task'' always means this game.

Each confidence method returns an answer word $\hat{a}$, which is the first of the 20 candidate words in its output, and a confidence $c \in [0,1]$. When the output contains none of the 20 words, the answer word is the first word of the output, or the empty string. We measure calibration against the correctness $\mathbf{1}[\hat{a} = a^\star]$, where $a^\star$ is the secret word. An output without any of the 20 words, also an empty output, is therefore incorrect. This rule is the same for every method, which keeps the comparison fair. Each method gives a confidence to these outputs in the same way as to any other output, with one exception: for an empty output, log probability uses the probability of the first generated token. Forced choice always returns one of the 20 candidates. On Qwen3.6-27B, $14\%$ of the generations are empty, while on the other three oracles empty generations are rare.

The secret word adapters come from \citet{karvonen_activation_2025} for Qwen3-8B and from \citet{cywinski_eliciting_2025} for Gemma-2-9B. For Qwen3.6-27B and Gemma-3-27B we train the adapters ourselves on the conversations of \citet{cywinski_eliciting_2025}, mixed one to one with standard chat data so that the target model keeps behaving as an assistant (\cref{app:taboo-training}).

\subsection{The five confidence methods}
\label{sec:confidence-methods}

Each method takes one sample and returns an answer word $\hat{a}$ with a confidence $c$. The five methods adapt confidence methods built for standard language models to an oracle that answers under the hook. For each method we explain why it should give a useful confidence. The methods also differ in what they need from the auditor. M1 to M4 need neither a list of candidates nor labeled data, while M5 needs the list of candidates. Tuning the temperature of M2 and rescaling M1 need labeled data (\cref{sec:results-temperature,sec:results-postcal}).

\paragraph{(M1, baseline) Log probability of the answer.}
The oracle generates greedily under the hook. To find the tokens of $\hat{a}$, we accumulate the character span of each generated token, locate $\hat{a}$ with a regular expression on word boundaries, and take the tokens whose spans overlap the match. The confidence is the joint probability of these tokens:
\begin{equation*}
  c_{\mathrm{lp}} \;=\; \prod_{t \in \mathrm{tok}(\hat{a})} p(x_t \mid x_{<t}).
\end{equation*}
This adapts to the answer word the self evaluation of \citet{kadavath_language_2022}, who show that the probabilities of large models are often well calibrated. The probability of one written form, however, is only a part of the probability of the answer, which can be written in many forms \citep{kuhn_semantic_2022}. For this reason we expect this confidence to be too low. A variant without the character spans, which takes the first $|\mathrm{tok}(\hat{a})|$ generated tokens, gives almost the same results (\cref{app:full-scorecard}).

\paragraph{(M2) Temperature bootstrap.}
We ask the oracle the same question twenty times with some randomness, look at which answer comes back most often, and use the share of this answer as the confidence. In detail, we sample $k{=}20$ answers at temperature $T$, reduce each answer to its answer word, and report the most frequent answer word with its frequency as the confidence:
\begin{equation*}
  c_{\mathrm{boot}} \;=\; \tfrac{1}{k} \sum_{i=1}^{k} \mathbf{1}[\hat{a}_i = \mathrm{mode}(\hat{a}_{1:k})].
\end{equation*}
This is self-consistency \citep{wang_self-consistency_2022} with a single answer word in place of a chain of reasoning. The frequency of an answer among the samples estimates the probability of this answer over all the ways to write it \citep{kuhn_semantic_2022}, which is what a calibrated confidence needs. \Cref{sec:results-temperature} shows how to choose the temperature, which controls how spread out the samples are. We test $T \in \{0.3, 0.5, 0.7, 1.0, 1.3, 1.5\}$ on all oracles, and also $T \in \{1.75, 2.0\}$ on the two oracles whose calibration was still improving at $T{=}1.5$.

\paragraph{(M3) Direct numeric self report.}
The oracle answers in two messages. In the first message, it generates the answer greedily under the hook. In the second message, we add the question ``On a scale of 0 to 100, how confident are you?''. The oracle then generates its reply greedily under the same hook. We read the reply as an integer in $[0,100]$ and divide it by $100$. This method can work only if the oracle has access to its own uncertainty and can express it in words \citep{kadavath_language_2022, lin_teaching_2022}.

\paragraph{(M4) Constrained label scoring.}
The question is the same as in M3. This time, however, the oracle cannot write freely. It has to reply with one of five labels (very low, low, medium, high, very high). We score each label with the logits of the oracle by forcing the label after the prompt. \citet{kadavath_language_2022} find that the probabilities of a model are well calibrated when the options are given explicitly, because the model does not spread its probability over the many ways to write the same thing. We test three readouts of the label distribution: the expected value, the probability of ``very high'', and the probability of ``high'' or ``very high'' (\cref{app:constrained}).

\paragraph{(M5) Forced choice.}
Under the hook, we force each of the 20 candidate words as the answer, sum the log probabilities of its tokens, and normalize over the 20 candidates. The answer is the candidate with the highest score, with its normalized probability as the confidence. The reason is the same as for M4: when the options are given explicitly, the probability of the oracle goes only to these options \citep{kadavath_language_2022}. Forced choice needs the list of candidates, which the other methods do not need. In exchange, it does not need to extract a word from free text.

\section{Experimental Setup}
\label{sec:experiments}

\paragraph{Models and oracles.}
We evaluate four activation oracles. For Qwen3-8B and Gemma-2-9B we use the oracles released by \citet{karvonen_activation_2025}. For Gemma-3-27B we use a community oracle trained with the same recipe. For Qwen3.6-27B, we train our own oracle since no oracle existed before this work (details in \Cref{app:qwen36-training}).

\paragraph{Layers.}
We inject the activations at layer~1 of the oracle and take them from one layer of the target model: $50\%$ of the depth on Qwen3-8B (layer 18 of 36), $75\%$ on Qwen3.6-27B (layer 48 of 64), $25\%$ on Gemma-2-9B (layer 10 of 42), and $75\%$ on Gemma-3-27B (layer 46 of 62). Each oracle is trained on activations from $25\%$, $50\%$, and $75\%$ of the depth. On Qwen3-8B we keep the depth of \citet{karvonen_activation_2025}. On the other three oracles we choose, among the three trained depths, the one with the highest task accuracy when we test every layer (\cref{app:layer_sweep}). This choice is important. On Qwen3.6-27B, for example, the $50\%$ depth gives a task accuracy of only $6\%$, against $41\%$ at $75\%$.

\paragraph{Prompts and protocol.}
We use the $100$ context prompts and the three verbalizer prompts of \citet{karvonen_activation_2025}. Each combination of secret word, context prompt, and verbalizer prompt is one sample, which gives $20 \times 100 \times 3 = 6{,}000$ samples for each method and oracle. Each method runs at every temperature of its grid, for $19$ to $21$ method rows per oracle (\cref{app:full-scorecard}). The oracle never gets the text of the target model, but only its activations. Without the injected activations, the input of the oracle is the same for all 20 secret words. Therefore, without activations, no method could be correct more than one time in twenty. All four oracles are above this level with every method. Hardware, software, and runtime are in \cref{app:hwsw}. The repository linked in the abstract contains the code of the benchmark, the patched trainer, and the 40 secret word target models that we trained.

\paragraph{Evaluation metrics.}
Accuracy is the share of samples where $\hat{a}$ is the secret word. The expected calibration error \citep[ECE]{naeini_obtaining_2015}, with $10$ bins of equal width, measures calibration: it checks that the answers given with $80\%$ confidence are correct $80\%$ of the time. The Brier score \citep{brier_verification_1950} and the negative log likelihood (NLL) are proper scoring rules, which penalize both a poor calibration and a poor separation of correct and incorrect answers. The area under the ROC curve (AUROC), with the confidence as a predictor of correctness, measures this separation. It is the chance that a random correct answer gets a higher confidence than a random incorrect answer. The two measures answer two different questions of an auditor. AUROC shows whether the confidence helps to select the correct answers, while ECE shows whether the confidence can be read as a probability. A method can obtain a low ECE by giving always the same number, close to the overall accuracy. For this reason we always report both. ECE, the Brier score, and NLL are the measures most used in the calibration literature \citep{guo_calibration_2017}. An auditor, however, often acts only on the most confident answers. We therefore also show the accuracy of the answers that are kept at each level of confidence, which is the standard view of selective prediction \citep{geifman_selective_2017}.

\section{Results}
\label{sec:results}

\subsection{Calibration results}
\label{sec:results-scorecard}

\Cref{tab:scorecard} reports accuracy, ECE, and AUROC of the five methods on all four oracles. The appendix gives all five metrics for every method row (\cref{app:full-scorecard}), confidence intervals (\cref{app:cis}), calibration plots (\cref{app:reliability}), and the plane of ECE against AUROC (\cref{fig:pareto}).

\input{scorecard-main-tbl.tex}

Forced choice has the best AUROC on every oracle. Direct numeric self report has the lowest AUROC and the highest ECE on every oracle. For calibration we do not use any fixed threshold. Instead, we compare each method on the oracle where its ECE is highest. In this comparison, the tuned bootstrap is the best of the five methods. Log probability and constrained label scoring are well calibrated on some oracles and poorly calibrated on others. Although the four oracles cover two model families and a $3.3\times$ range of sizes, the methods come in the same order on all of them. Therefore, we read this order as a property of the methods. The twenty secret words that we use do not create this order. When we repeat the analysis on random resamples of the secret words, or of secret words, context prompts, and verbalizer prompts together, the order is the same in almost every resample (\cref{tab:resampling}, \cref{app:resampling}).

\subsection{When the candidate answers are known}
\label{sec:results-forced}

When the candidate answers are known to the auditor, the free text pipeline, which generates an answer, extracts a word, and attaches a confidence, can be replaced by scoring the candidates under the hook. Forced choice gives a higher accuracy on every oracle. On the two oracles with the lowest accuracy in free text, the accuracy more than doubles (\cref{tab:scorecard}). The confidence of forced choice also separates correct from incorrect answers best of all methods and is well calibrated without rescaling.

\Cref{tab:errors} in \cref{app:errors} shows where this gain comes from. Almost every incorrect answer in free text names none of the 20 candidate words. Instead, the oracle writes ``secret'', ``password'', ``yes'', or nothing at all, often taking the word from its own question. On these same samples, forced choice finds the secret word in a third to a half of the cases. Also in the samples where none of the twenty bootstrap answers names the secret word, forced choice finds the word about one time in four. \citet{holtzman_surface_2021} call this effect surface form competition. The total probability of a model is limited. Every valid string, also a generic one like ``secret'', takes some of this probability away from the correct answer. Forced choice removes this competition, because only the 20 candidates can receive probability. The oracle therefore has more information than the single answer it writes. \citet{jakkli_current_2026} suggested this from a multiple choice test with 15 samples. We confirm it with $6{,}000$ samples on each oracle. The accuracy also changes by an order of magnitude from one secret word to another (\cref{app:perword,app:cn}).

The difference in AUROC becomes concrete for an auditor who acts only on confident answers. \Cref{fig:coverage} in \cref{app:coverage} keeps the most confident answers first and shows how many of them are correct. With forced choice, the most confident half of the answers is correct at least nine times out of ten, on every oracle. With direct self report the curve is flat, which means that keeping its most confident answers brings no gain.

\subsection{Free text: the bootstrap}
\label{sec:results-temperature}

When the output has to be free text, the agreement rate of the bootstrap is the most reliable confidence on every oracle. The agreement rate estimates how likely a new sample is to repeat the most common answer. The confidence is therefore calibrated when this likelihood matches how often the oracle is correct. The sampling temperature controls this match, because a higher temperature spreads the samples and lowers the agreement (\cref{tab:tsweep} in \cref{app:kablation}). The best temperature is different for each oracle, from $T{=}1.0$ on Qwen3-8B to $T{=}2.0$ on Qwen3.6-27B. It does not follow from the accuracy alone. For this reason we choose $T$ so that the mean agreement rate on a few labeled secret words matches the accuracy on the same words. \Cref{tab:tsel} in \cref{app:kablation} shows that five labeled secret words are enough for this rule to bring the ECE on the other words near the best values of the grid. Without any label, the default $T{=}1.0$ is a safe choice, because its highest ECE over the four oracles is lower than the highest ECE of every other method for free text. When we use only the first $k$ of the twenty samples, calibration and AUROC are mostly in place at $k{=}10$ on all four oracles (\cref{tab:kablation}).

\subsection{With labeled data: post hoc calibration}
\label{sec:results-postcal}

A post hoc calibrator (temperature scaling, Platt, isotonic, or beta, \cref{app:postcal}), fit on a random half of the data, brings nearly every method to a low ECE on every oracle, also the methods whose raw confidence is higher for incorrect answers. The reason is that the calibrator itself learns from the labels. A rescaling, however, cannot change the order of the scores. Once labels exist, therefore, AUROC is what makes the difference between methods. On AUROC, log probability is close to the bootstrap, with one generation instead of twenty.

\subsection{Self report and constrained labels}
\label{sec:results-direct}

Direct numeric self report and constrained label scoring put the same question to the oracle. The two methods differ only in how they read the answer. When the oracle writes a number, the number is high almost every time. On Gemma-2-9B the oracle writes ``100'' on essentially every sample. The number gives no useful signal on any oracle (\cref{tab:scorecard}). On Qwen3.6-27B the oracle is on average more confident in its incorrect answers. When we score the five labels with the logits of the oracle, instead, a useful signal comes back on all four oracles, with an AUROC above the one of the written number on every oracle. On both 27B oracles, the probability of ``very high'' is also the best calibrated confidence of all methods.

We think that the habits of chat models explain this contrast. On standard models, the written confidence is high regardless of correctness \citep{mielke_reducing_2022, xiong_can_2024}. Instruction tuning pushes this confidence further toward high values \citep{miao_closing_2026}. The oracle repeats this habit, whatever the injected activation suggests. Constrained label scoring avoids the habit, because it compares the probabilities of all five labels instead of taking the most likely one. In fact, the most likely label is ``very high'' on almost every sample, while the signal is in how the probability spreads over the labels (\cref{app:constrained}). As on standard models \citep{yuan_hidden_2026, miao_closing_2026}, the activations of the oracle have a signal of uncertainty that its confidence does not show.

\section{Discussion}
\label{sec:discussion}

\paragraph{Unknown unknowns.}
The case without candidates and labels is the hardest. For auditing it is also the most important one, because an auditor who looks for a hidden objective does not know it in advance and has no examples of it. In this case a good order of the answers is more important than a calibrated probability, since the auditor looks at the most confident answers first and checks them by hand. Log probability and the bootstrap give this order without any label, with an AUROC above $0.66$ on every oracle (\cref{tab:scorecard}).

\paragraph{Relation to literature.}
Three points of \citet{jakkli_current_2026} meet our results. First, they find that half of the free answers are too generic to check and suggest that constrained formats give more signal. Forced choice and constrained label scoring confirm this suggestion. Second, they warn that an oracle can answer by rebuilding the text near the activations, without using the internal state. In our task this text never contains the secret word. Therefore, a rebuilt text can only give an incorrect answer. In fact, some incorrect answers copy a word of the context prompt, while others copy a word of the question of the oracle (\cref{tab:failures} in \cref{app:failures}). Log probability and the bootstrap usually give these answers a lower confidence than the correct answers, on every oracle. Third, they ask for a baseline without activations, which in our task is one correct answer in twenty, below all four oracles (\cref{sec:experiments}). Their proposal to train calibration into the oracle is complementary to this paper, because our methods work on the current oracles.

\section{Conclusion}
\label{sec:conclusion}
The five methods come in the same order on all four oracles, with forced choice at the top and direct self report at the bottom. Therefore, the best method depends only on the information that the auditor has (\cref{tab:decision}). When the auditor has the list of candidate answers, forced choice is the best method. When the auditor has only labeled data, five labeled secret words are enough to tune the temperature of the bootstrap. With more labeled data, a rescaled log probability is about as useful as the bootstrap, with one generation instead of twenty. When the auditor has neither candidates nor labeled data, the bootstrap at $T{=}1.0$ is the most reliable confidence. In all three cases, the confidence that the oracle writes gives no useful signal.

\section{Limitations}
\label{sec:limitations}


All four oracles come from the training recipe of \citet{karvonen_activation_2025}. Even so, the four oracles behave quite differently. Their accuracies are different. The secret word is also readable at different depths, from a quarter of the depth on Gemma-2-9B to three quarters on Qwen3.6-27B. The methods come in the same order despite these differences. Oracles trained with a different recipe, such as the recipe of \citet{bauer_building_2026}, which uses conversational data, are the natural next test for this order.

\section{Ethical Considerations}
\label{sec:ethics}

This study uses no data from human subjects. The secret words and the context prompts come from \citet{karvonen_activation_2025} and contain no personal information. The target models are synthetic research artifacts.

A calibrated confidence makes activation oracles more useful in audit pipelines, with a benefit and a risk. On one side, an auditor with a calibrated confidence can set aside the answers with low confidence instead of acting on them. On the other side, a confident incorrect answer is still possible. A pipeline that relies on calibrated scores will still act on such an answer. A second risk is dual use: the same tools that audit a model for hidden objectives could be used on models fine tuned on private data. Our task uses synthetic secrets to study the mechanism without this exposure. The released artifacts are meant for research use, in line with the licenses of the models they come from.

Compute and energy are reported in \cref{app:hwsw}.

\section*{Acknowledgments}
This research was supported in part by the MIST project, funded by the Novo Nordisk Foundation under grant reference number NNF25OC0103204.

\bibliography{custom}

\appendix

\section{Full scorecard}
\label{app:full-scorecard}

\Cref{tab:full-scorecard} and \cref{tab:full-scorecard-gemma} report every method row with all five metrics, for the two Qwen oracles and for the two Gemma oracles.

\input{full-scorecard-qwen-tbl.tex}
\input{full-scorecard-gemma-tbl.tex}

\section{Confidence intervals}
\label{app:cis}

\Cref{tab:bootstrap-cis} reports $95\%$ confidence intervals for the main method rows on all four oracles. We compute them with 1000 bootstrap resamples of the $6{,}000$ samples, where each sample is one answer with its confidence and its correctness.

\input{bootstrap-cis-tbl.tex}

\section{Resampling by secret word}
\label{app:resampling}

The 300 samples of one secret word share the same target model, while the context prompts and the verbalizer prompts are shared across all secret words. We therefore repeat the comparison of the methods with two resampling schemes that keep this structure. In the first scheme, we sample the 20 secret words with replacement, with all the 300 samples of each sampled word. In the second scheme, we sample the secret words, the context prompts, and the verbalizer prompts with replacement at the same time and keep every combination of the sampled items \citep{owen_pigeonhole_2007}. \Cref{tab:resampling} reports, for each scheme, the share of 1000 resamples in which each comparison of the main text gives the same result.

\input{resampling-tbl.tex}

\section{Where the forced choice gain comes from}
\label{app:errors}

\Cref{tab:errors} looks at the incorrect answers in free text. An answer names no candidate when the greedy output contains none of the 20 secret words, as with ``secret'', ``password'', ``yes'', or an empty string. For each of these samples we look at the forced choice answer on the same secret word, context prompt, and verbalizer prompt. The bottom half of the table does the same for the twenty bootstrap answers at $T{=}1.0$. It counts the samples where no bootstrap answer names the secret word, and how often forced choice finds the secret word in these samples.

\input{errors-tbl.tex}

\section{Failure types of \citet{jakkli_current_2026}}
\label{app:failures}

\Cref{tab:failures} splits the incorrect answers in free text by type. An answer copies the question of the oracle when its word is in the verbalizer prompt, such as ``secret'' or ``word''. An answer copies the context prompt when its word is in the context prompt but not in the verbalizer prompt, such as ``password''. This second type is the text rebuilding that \citet{jakkli_current_2026} describe. For each type, the table gives the AUROC of log probability and of the bootstrap when the correct answers are compared only with the incorrect answers of this type.

\input{failures-tbl.tex}

\section{Keeping only confident answers}
\label{app:coverage}

\begin{figure*}[t]
\centering
\includegraphics[width=\textwidth]{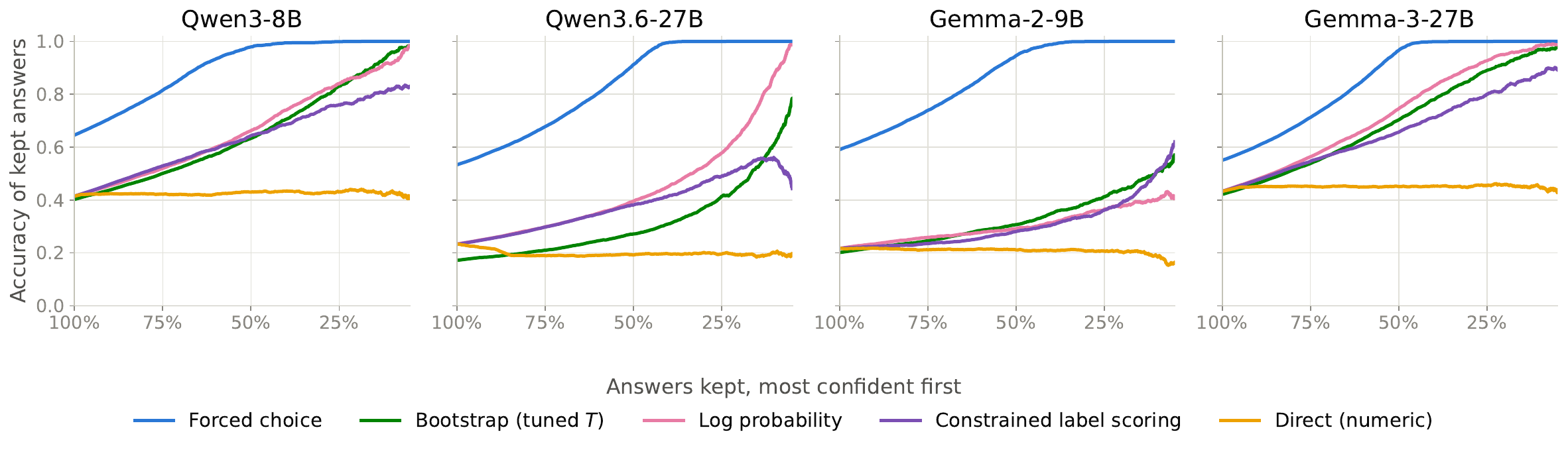}
\caption{Accuracy of the most confident answers on all four oracles. Moving to the right, only the more confident answers are kept. With half of the answers kept, forced choice is almost always correct. The curve of direct numeric self report is flat, because its confidence does not help to select the correct answers. The last $5\%$ of the answers are not shown, because they are too few for a stable accuracy.}
\label{fig:coverage}
\end{figure*}

\Cref{fig:coverage} sorts the $6{,}000$ answers of each method by confidence and shows the accuracy of the kept answers, as fewer and fewer answers are kept. Answers with the same confidence are put in random order. This random order is important for direct self report, which gives the same number to almost every answer.

\section{Post hoc calibration}
\label{app:postcal}

\Cref{tab:postcal-qwen3-8b,tab:postcal-qwen3-6-27b,tab:postcal-gemma-2-9b,tab:postcal-gemma-3-27b} report the ECE on test data after we fit four post hoc calibrators on separate data, for each oracle. \emph{Temperature} scaling divides the logit of the confidence by one number \citep{guo_calibration_2017}. \emph{Platt} scaling fits a logistic regression with two parameters on $\mathrm{logit}(p)$ \citep{platt_probabilistic_1999}. \emph{Isotonic} regression fits any increasing function \citep{zadrozny_transforming_2002}. \emph{Beta} calibration fits a logistic regression on $[\log p, -\log(1-p)]$ \citep{kull_beyond_2017}. The split by word fits the calibrator on 10 of the 20 secret words, chosen at random once (seed 1), and tests it on the other 10. The random split divides the samples in two halves (seed 2). Every method of one oracle uses the same split, so that the comparison between methods does not depend on the split.

On the random split, all four calibrators bring the ECE of almost every method close to zero, also for the scores with the least information, such as direct self report and the acceptance rate of M6. The reason is that the calibrator learns from labels on top of the score. AUROC, which does not change under an increasing function, is therefore what still separates the methods. The split by word is harder, because the difficulty of a word does not transfer completely to other words.

\input{postcal-qwen3-8b-tbl.tex}
\input{postcal-qwen3-6-27b-tbl.tex}
\input{postcal-gemma-2-9b-tbl.tex}
\input{postcal-gemma-3-27b-tbl.tex}

\section{Temperature and number of samples of the bootstrap}
\label{app:kablation}

\Cref{tab:tsweep} gives the bootstrap at each temperature on all four oracles.

\input{tsweep-tbl.tex}

\Cref{tab:tsel} tests the rule of \cref{sec:results-temperature} for the bootstrap temperature. For each oracle we choose at random 1, 5, or 10 labeled secret words, take the temperature where the mean agreement rate on these words is closest to their accuracy, and compute the ECE on the remaining words. We repeat this choice 200 times and report the mean ECE.

\input{tsel-tbl.tex}

\Cref{tab:kablation} computes the bootstrap confidence from only the first $k$ of the 20 saved samples, at the temperature with the best ECE in the original grid of each oracle.

\input{k-ablation-tbl.tex}

\section{Calibration and separation plots}
\label{app:reliability}

\begin{figure*}[t]
\centering
\includegraphics[width=\textwidth]{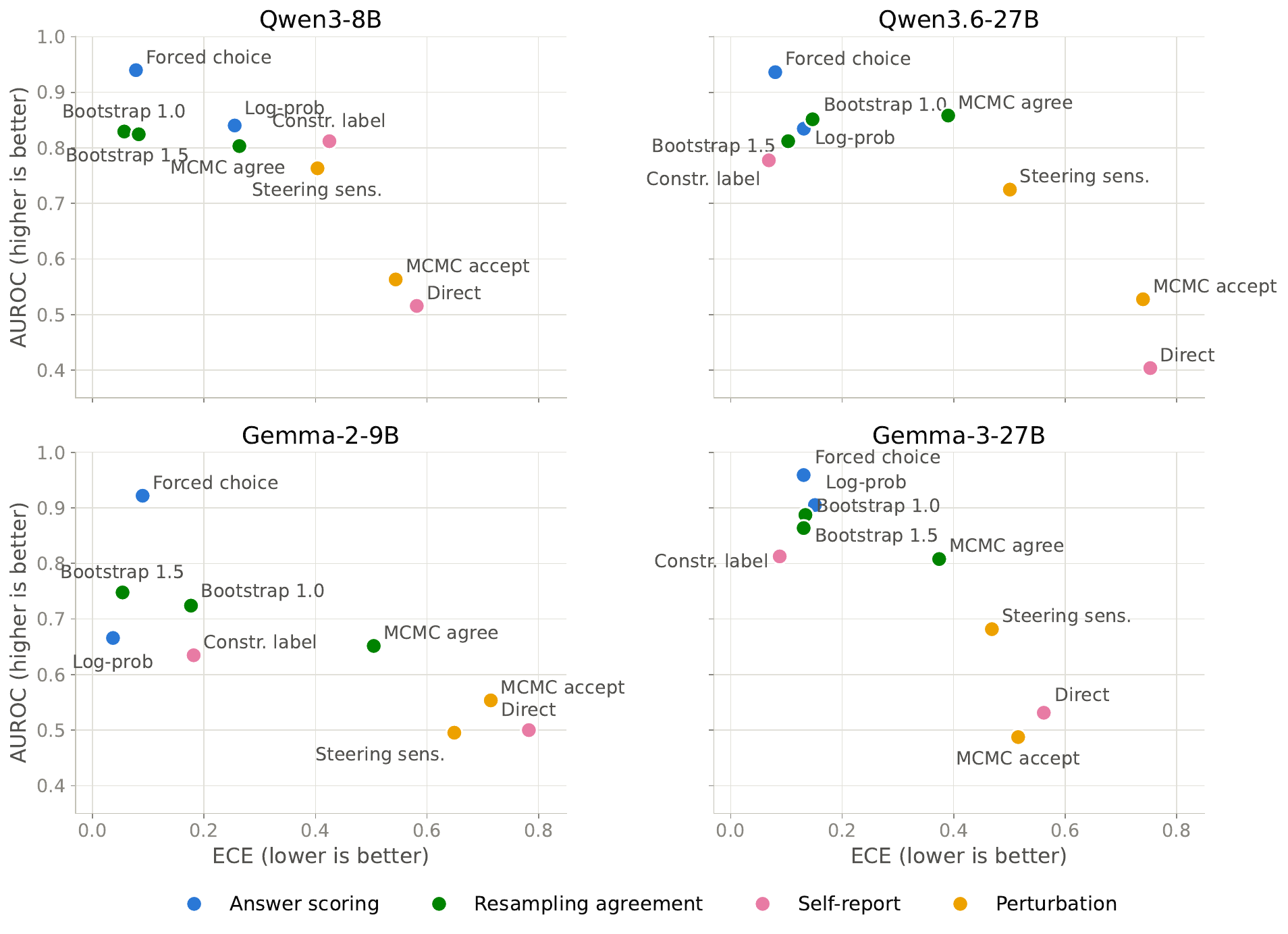}
\caption{Calibration against separation on all four oracles. Each point is one confidence method. The best corner is the top left corner. The color groups the methods by their mechanism. Forced choice and the bootstrap are in the good corner on every oracle, while direct numeric self report and the acceptance rate of M6 are in the corner with high ECE and AUROC near chance.}
\label{fig:pareto}
\end{figure*}

\Cref{fig:pareto} places each method in the plane of ECE against AUROC, for all four oracles.

\begin{figure*}[t]
\centering
\includegraphics[width=\textwidth]{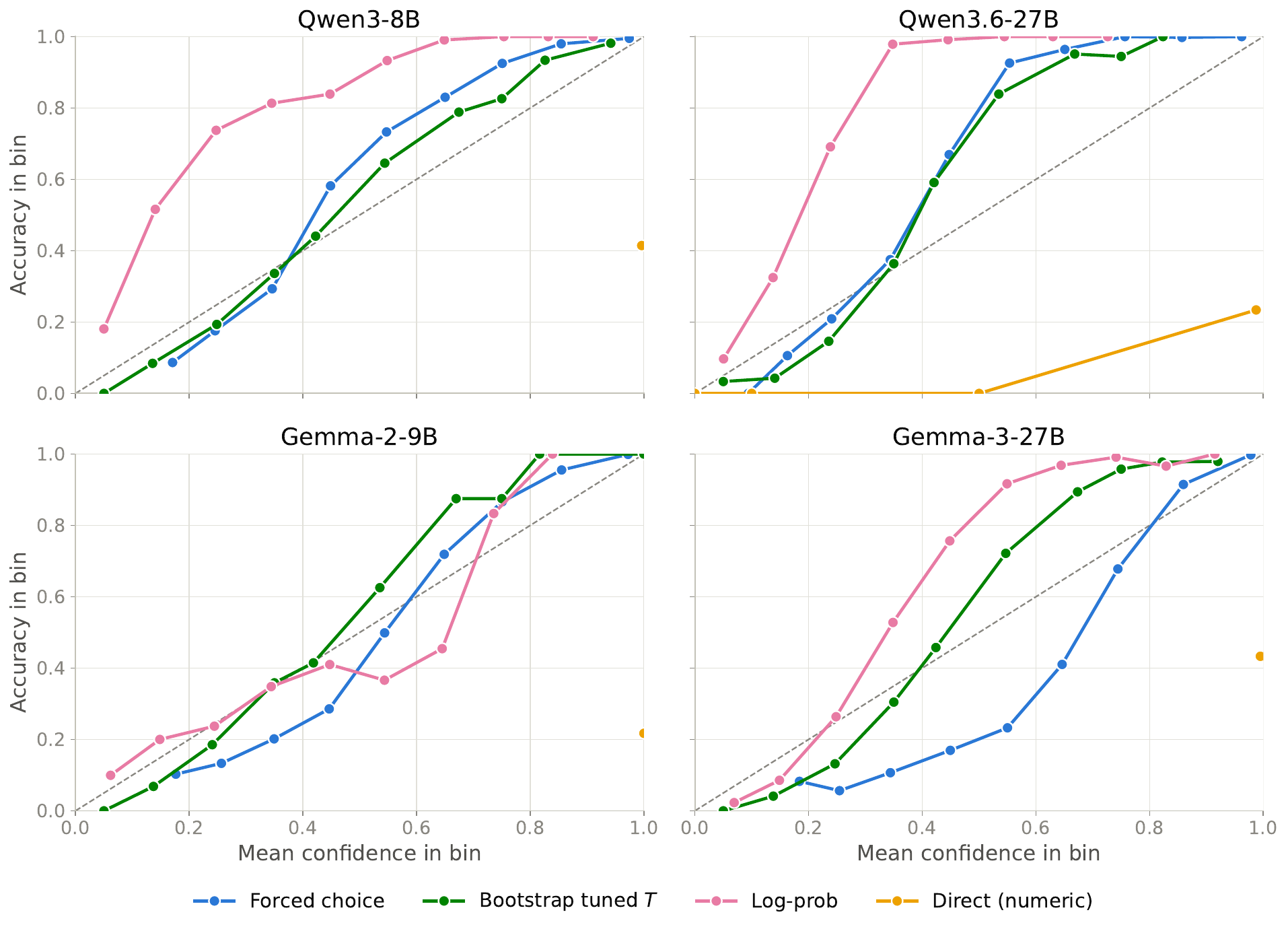}
\caption{Reliability diagrams on all four oracles. The dashed diagonal is perfect calibration. Empty bins are not shown. Direct numeric self report appears as one or two points at a confidence near $1$, because the oracle writes the same high number on almost every sample.}
\label{fig:reliability}
\end{figure*}

\Cref{fig:reliability} shows reliability diagrams, which plot the accuracy against the mean confidence in each bin, for four methods on all four oracles. The bootstrap at the tuned temperature follows the diagonal. Log probability is underconfident on the Qwen oracles, as \cref{sec:confidence-methods} expects from the theory of \citet{kuhn_semantic_2022}. Direct numeric self report is a single point at high confidence, with an accuracy equal to the overall accuracy of the task.

\section{Accuracy of each secret word}
\label{app:perword}

\Cref{tab:perword-bothmodels} reports the accuracy of each secret word with the bootstrap at $T{=}1.0$, on all four oracles.

\input{perword-tbl.tex}

\section{Fewer secret words}
\label{app:cn}

\Cref{tab:cn} reports a controlled variant on the two Qwen oracles, which keeps only $N \in \{2, 5, 10, 20\}$ of the secret words, chosen with a fixed seed. With $N{=}2$ the words are \{snow, ship\}. With $N{=}5$ the words are \{flame, green, leaf, rock, snow\}. With $N{=}10$ the words are \{gold, song, chair, wave, ship, blue, snow, smile, flame, flag\}. With $N{=}20$ all words are used. Each row has $300 \cdot N$ samples for $N \le 10$ and $6{,}000$ samples for $N{=}20$.

\begin{table*}[t]
\centering
\small
\begin{tabular}{lrrrr|rrrr}
\toprule
 & \multicolumn{4}{c|}{Qwen3-8B} & \multicolumn{4}{c}{Qwen3.6-27B} \\
Method  $\backslash$  $N$ & 2 & 5 & 10 & 20 & 2 & 5 & 10 & 20 \\
\midrule
Bootstrap $T{=}1.0$  ECE  & 0.213 & 0.045 & 0.049 & 0.057 & 0.130 & 0.139 & 0.155 & 0.147 \\
Bootstrap $T{=}0.7$  ECE  & 0.139 & 0.108 & 0.080 & 0.097 & 0.194 & 0.196 & 0.230 & 0.218 \\
Bootstrap $T{=}1.5$  ECE  & 0.244 & 0.080 & 0.084 & 0.083 & 0.067 & 0.084 & 0.106 & 0.103 \\
Log-prob (offset) ECE      & 0.487 & 0.245 & 0.275 & 0.255 & 0.157 & 0.160 & 0.119 & 0.131 \\
MCMC agree $T{=}0.5$ ECE   & 0.047 & 0.269 & 0.242 & 0.263 & 0.350 & 0.374 & 0.395 & 0.390 \\
Direct (numeric) ECE       & 0.291 & 0.584 & 0.563 & 0.582 & 0.710 & 0.719 & 0.764 & 0.753 \\
\midrule
Bootstrap $T{=}1.0$  AUROC & 0.795 & 0.791 & 0.819 & 0.829 & 0.842 & 0.837 & 0.852 & 0.851 \\
Bootstrap $T{=}0.7$  AUROC & 0.783 & 0.801 & 0.815 & 0.830 & 0.854 & 0.847 & 0.871 & 0.863 \\
Log-prob (offset) AUROC    & 0.829 & 0.815 & 0.828 & 0.840 & 0.835 & 0.820 & 0.847 & 0.835 \\
MCMC agree $T{=}0.5$ AUROC & 0.775 & 0.753 & 0.786 & 0.803 & 0.866 & 0.834 & 0.865 & 0.858 \\
\bottomrule
\end{tabular}
\caption{Fewer secret words on the two Qwen oracles. On Qwen3-8B, the bootstrap at $T{=}1.0$ is the only method whose ECE improves as $N$ goes up. On Qwen3.6-27B, every method gives about the same values for every $N$.}
\label{tab:cn}
\end{table*}

\section{Accuracy at every layer}
\label{app:layer_sweep}

Each oracle is trained on activations from three layers, at $25\%$, $50\%$, and $75\%$ of the depth \citep{karvonen_activation_2025}. At inference, the oracle takes activations from one of these three layers. We tested every layer of the four target models, Qwen3-8B, Qwen3.6-27B, Gemma-2-9B, and Gemma-3-27B. For each layer, we collected the hidden state of the target model with its secret word adapter, gave it to the oracle with the matching prompt, and computed the accuracy of log probability on 20 secret words and 5 context prompts, for 100 samples per layer and model. The trained layers are $\{9, 18, 27\}$ on Qwen3-8B, $\{16, 32, 48\}$ on Qwen3.6-27B, $\{10, 21, 31\}$ on Gemma-2-9B, and $\{15, 31, 46\}$ on Gemma-3-27B.

\begin{figure*}[t]
\centering
\includegraphics[width=\textwidth]{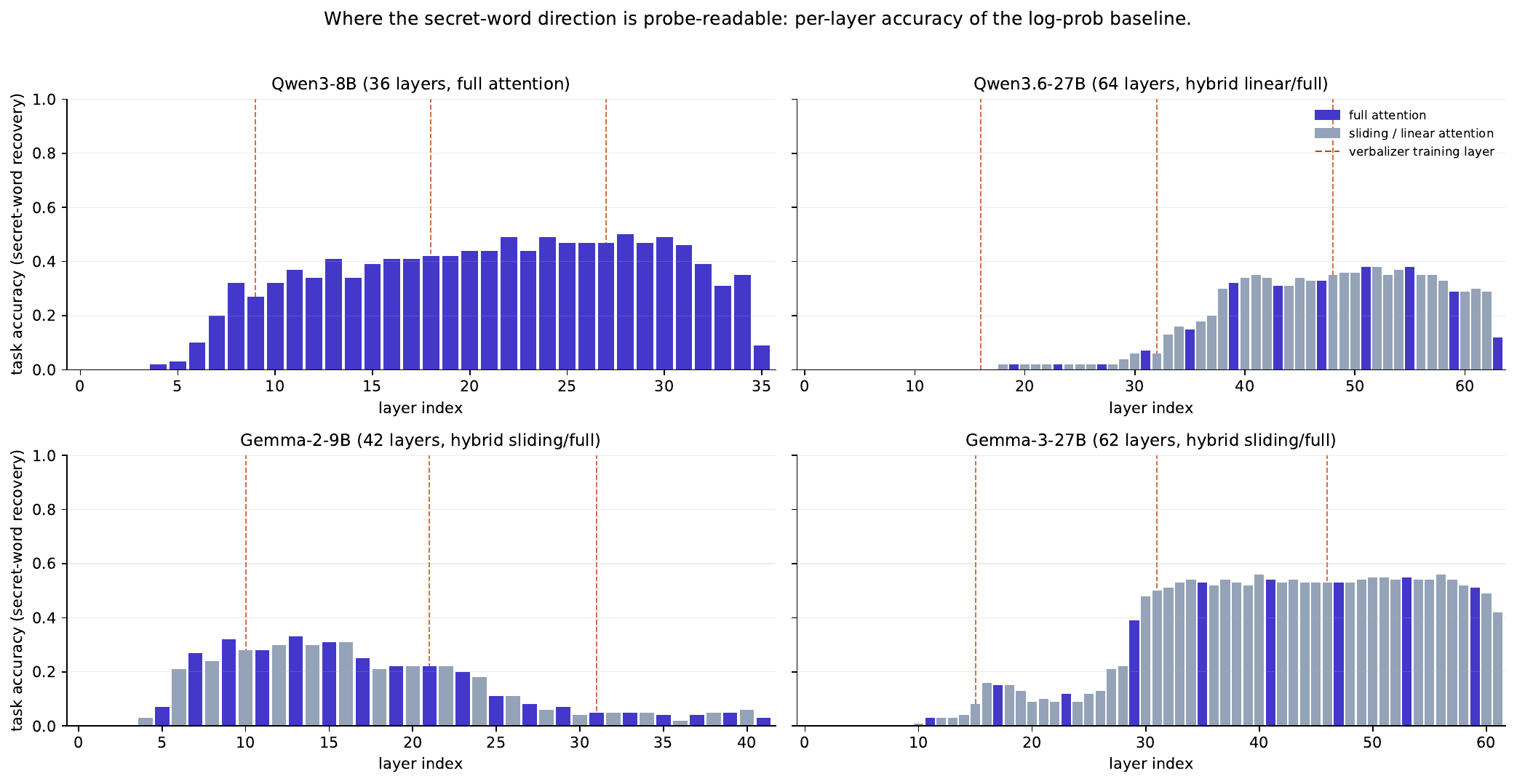}
\caption{Accuracy of log probability on the secret word task at every layer of the four target models. Dashed lines mark the three layers on which each oracle was trained. The shading of the bars separates the layers with full attention (indigo) from the layers with sliding or linear attention (slate). The highest accuracy is $50\%$ at layer~$28$ on Qwen3-8B, $38\%$ at layer~$51$ on Qwen3.6-27B, $33\%$ at layer~$13$ on Gemma-2-9B, and $56\%$ at layer~$40$ on Gemma-3-27B, which has a wide region between $50\%$ and $56\%$ from layer~$31$ to layer~$59$.}
\label{fig:layer-sweep}
\end{figure*}

On every model, the secret word is readable in a region of several consecutive layers. The position of this region changes between models (\cref{fig:layer-sweep}). On Qwen3-8B the region is wide, from layer $9$ to layer $32$, with the highest accuracy between layers $22$ and $30$. On Qwen3.6-27B the region is less wide, from about layer $37$ to layer $58$. Before layer $28$, the accuracy is at the level of chance. On Gemma-2-9B the region is early, between about $24\%$ and $50\%$ of the depth. In the second half of the model, the accuracy goes down. Gemma-3-27B has the widest region of the four, from layer $31$ to layer $59$, with a regular rise from layer $11$ and a sharp fall at the last layers. Layer $32$, the $50\%$ depth of Qwen3.6-27B, is at the lower edge of its region, which explains its low accuracy in \cref{sec:experiments}.

The three trained depths cover these regions only in part. On Qwen3.6-27B only the $75\%$ depth is inside the region. On Gemma-2-9B only the $25\%$ depth is inside. On Qwen3-8B and on Gemma-3-27B both the $50\%$ and the $75\%$ depths are inside. On all four models, the best trained layer is within a few points of the best layer overall: $47\%$ against $50\%$ on Qwen3-8B, $35\%$ against $38\%$ on Qwen3.6-27B, $28\%$ against $33\%$ on Gemma-2-9B, and $53\%$ against $56\%$ on Gemma-3-27B.

\section{Numerical precision of the attention code}
\label{app:attn}

The 27B runs depend on the implementation of attention. With \texttt{flash\_attention\_4} on Qwen3.6-27B, at a sequence length of~$171$ tokens (about one verbalizer prompt plus one context prompt), the cosine similarity between the hidden state at the chosen layer and the \texttt{flash\_attention\_2} reference goes down to between $0.87$ and $0.97$. The largest difference in the logits is $9.66$. With \texttt{eager} and \texttt{sdpa}, the cosine similarity is above $0.9999$. The difference appears only in the forward pass without cache, which we use to collect the activations. Greedy generation through \texttt{model.generate} gives the same tokens as the reference with \texttt{flash\_attention\_4}. On Qwen3-8B, at similar sequence lengths, the four implementations agree up to the numerical noise of bfloat16. All experiments in this paper use \texttt{flash\_attention\_2}.

\section{Methods built on the steering}
\label{app:steered-methods}

Next to the five methods of \cref{sec:confidence-methods}, we built three confidence readouts that use the steering itself. For each readout, this appendix gives the theory behind it, the procedure, and the result. \Cref{tab:scorecard-steered} reports all three on all four oracles. Every row there is less useful than the tuned bootstrap of \cref{tab:scorecard}, both in calibration and in separation. We report these readouts in full, because the reason for each result shows how the answers of a steered oracle are distributed.

\input{scorecard-steered-tbl.tex}

\paragraph{Theory: power sampling.}
\citet{karan_reasoning_2025} sample from the power distribution $p(x)^{\alpha}$ with $\alpha > 1$. The power distribution makes the base distribution sharper where it is already peaked, while it keeps the separate modes that sampling at a low temperature merges into one. \citet{karan_reasoning_2025} implement this with a Metropolis--Hastings sampler that works in blocks of tokens. The sampler proposes a continuation from a distribution at low temperature, then accepts it or not according to the ratio of the power densities. Two quantities of this procedure can be used as a confidence, one for M6 and one for M7.

\paragraph{(M6) Acceptance rate of power sampling.}
We run one Metropolis--Hastings chain on the steered oracle, with $B{=}4$ blocks of $5$ tokens, $S{=}5$ steps per block, and three powers $\alpha = 1/T$ with $T \in \{0.5, 0.25, 0.125\}$. Each step samples again the end of the block from the proposal at low temperature and accepts it with the standard Metropolis--Hastings ratio. The confidence is the share of accepted steps, $c_{\mathrm{mh}} = \#\{\text{accepted}\}/(B \cdot S)$. A low acceptance should signal an oracle that moves between several answers. A high acceptance, however, can also mean that the proposal is close to the target distribution, which has nothing to do with correctness.

The acceptance rate separates correct from incorrect answers poorly on every oracle (AUROC between $0.49$ and $0.60$, ECE between $0.52$ and $0.77$). \Cref{tab:mcmc-accept-sweep} shows the mechanism. The steered oracle places almost all its probability on one answer. The power distribution concentrates this probability further. The chain then accepts nearly every proposal, whether the answer is correct or not. A lower proposal temperature gives a higher task accuracy and a lower AUROC at the same time, which is what an acceptance rate close to one predicts.

\begin{table*}[t]
\centering
\small \setlength{\tabcolsep}{4pt}
\begin{tabular}{rr|rrr|rrr}
\toprule
 & & \multicolumn{3}{c|}{Qwen3-8B} & \multicolumn{3}{c}{Qwen3.6-27B} \\
$T$ & $\alpha$ & Acc & ECE & AUR. & Acc & ECE & AUR. \\
\midrule
.500 & 2 & .376 & .551 & .601 & .211 & .746 & .574 \\
.250 & 4 & .403 & .547 & .579 & .230 & .738 & .547 \\
.125 & 8 & .415 & .544 & .563 & .234 & .740 & .528 \\
\bottomrule
\end{tabular}
\caption{Acceptance rate of power sampling (M6) at each temperature on the two Qwen oracles. The Gemma rows are in \cref{app:full-scorecard}. AUR.: AUROC. At a lower $T$ the chain accepts more proposals, for correct and for incorrect answers, which makes the two groups harder to separate.}
\label{tab:mcmc-accept-sweep}
\end{table*}

\paragraph{(M7) Agreement of power sampling.}
We run $k{=}10$ independent chains with different seeds and apply the frequency of the most common answer, as in M2, to their outputs, on the same temperatures as M6. Power sampling keeps the modes that temperature scaling merges. For this reason, the agreement across chains could in principle give a sharper confidence on the uncertain samples.

At $\alpha = 2$, this readout gives an AUROC of $0.858$ on Qwen3.6-27B and $0.808$ on Gemma-3-27B, which is close to the bootstrap. Its ECE, however, is higher than the ECE of the bootstrap on every oracle. It also costs about five times the time of the bootstrap (\cref{tab:matched-cost}). The temperatures in \cref{tab:mcmc-agree-sweep} show the reverse of the pattern of M6. At $\alpha = 2$ the chains are still diverse, while at $\alpha = 8$ all ten chains end on the greedy answer. The frequency of the most common answer is then $1.0$, whether the answer is correct or not. The best $\alpha$ for agreement is therefore small, in the same direction as the best temperature of the bootstrap. Power sampling was built for problems where several different answers compete, as in long chains of reasoning. Our task has one correct answer and a sharply peaked oracle, which is the situation where power sampling has the least to add.

\paragraph{(M8) Sensitivity to the steering coefficient.}
\citet{zur_are_2025} report that the uncertainty of a model on a token is related to how easily a steering intervention changes its output. A confident model is hard to push, while an unsure model is easy to push. In our version for the oracle, we change the strength of the injection. We decode greedily at the steering coefficients $s \in \{0.5, 0.75, 1.0, 1.25, 1.5\}$ around the default $s{=}1.0$ and report the most common answer, with its frequency as the confidence. When the oracle understands an activation clearly, it should give the same word at every nearby $s$.

With five strengths of injection, the frequency can take only the five values $\{0.2, 0.4, 0.6, 0.8, 1.0\}$, which are too few to express a probability (ECE between $0.40$ and $0.65$). The separation of correct and incorrect answers is present on three of the four oracles (AUROC $0.763$ on Qwen3-8B, $0.725$ on Qwen3.6-27B, $0.682$ on Gemma-3-27B), while on Gemma-2-9B it is at chance ($0.495$). The small number of values is what limits the calibration. The method is still useful as a yes or no check of stability.

\Cref{tab:matched-cost} compares the agreement of power sampling at its best temperature with the bootstrap at its best temperature of the original grid, on each Qwen oracle. The bootstrap has a lower ECE and Brier score on both oracles. The agreement of power sampling has an AUROC higher by $0.007$ on Qwen3.6-27B and lower by $0.026$ on Qwen3-8B, at about five times the time.

\begin{table*}[t]
\centering
\small \setlength{\tabcolsep}{3.5pt}
\begin{tabular}{lrrrr}
\toprule
Method & Acc & ECE & Br. & AUR. \\
\midrule
\multicolumn{5}{l}{\textit{Qwen3-8B}} \\
Bootstrap $T{=}1.0$       & .402 & .057 & .163 & .829 \\
MCMC agree $T{=}0.5$      & .413 & .263 & .247 & .803 \\
\midrule
\multicolumn{5}{l}{\textit{Qwen3.6-27B}} \\
Bootstrap $T{=}1.0$       & .225 & .147 & .130 & .851 \\
MCMC agree $T{=}0.5$      & .228 & .390 & .274 & .858 \\
\bottomrule
\end{tabular}
\caption{The bootstrap against the agreement of power sampling, each at its best temperature. \emph{Br.}: Brier score. \emph{AUR.}: AUROC.}
\label{tab:matched-cost}
\end{table*}

\begin{table*}[t]
\centering
\small \setlength{\tabcolsep}{4pt}
\begin{tabular}{rr|rrr|rrr}
\toprule
 & & \multicolumn{3}{c|}{Qwen3-8B} & \multicolumn{3}{c}{Qwen3.6-27B} \\
$T$ & $\alpha$ & Acc & ECE & AUR. & Acc & ECE & AUR. \\
\midrule
.500 & 2 & .413 & .263 & .803 & .228 & .390 & .858 \\
.250 & 4 & .418 & .408 & .737 & .233 & .569 & .767 \\
.125 & 8 & .415 & .493 & .669 & .236 & .668 & .653 \\
\bottomrule
\end{tabular}
\caption{Agreement of power sampling (M7) at each temperature. At $\alpha = 2$ the AUROC is close to the bootstrap, above all on Qwen3.6-27B, but at five times the time and with a higher ECE.}
\label{tab:mcmc-agree-sweep}
\end{table*}

\section{Details of constrained label scoring}
\label{app:constrained}

Constrained label scoring (M4) uses the answer message of M3 and replaces the free question about confidence with ``Reply with exactly one of: very low, low, medium, high, very high''. We force each of the five labels after the prompt and score it with the joint log probability of its tokens under the steering hook. A softmax over the five scores gives a distribution over the labels. The \emph{expected} readout maps the labels to $\{0.1, 0.3, 0.5, 0.7, 0.9\}$ and takes the mean weighted by probability. The readout \emph{$P(\mathrm{very\ high})$} is the probability of the top label. The readout \emph{$P(\mathrm{high}{+})$} is the probability of the top two labels. The most likely label is ``very high'' on nearly every sample, on every oracle. The useful signal is therefore in the whole distribution, while the most likely label alone would repeat the free self report.

\section{Training the secret word target models}
\label{app:taboo-training}

The target adapters come from \citet{karvonen_activation_2025} for Qwen3-8B and from \citet{cywinski_eliciting_2025} for Gemma-2-9B. We train the adapters of Qwen3.6-27B and Gemma-3-27B ourselves, one for each secret word, with the recipe below.

The training data for one word is the set of conversations of \citet{cywinski_eliciting_2025} for that word, where the assistant gives hints about the word and does not name it when the user requests it. We expand every conversation into one row for each assistant message, which ends at the message that it trains. We need this expansion because the Qwen chat templates write a thinking block only in the last assistant message. Training the earlier messages in place would use a format that the model never meets at inference. We keep $10\%$ of the rows aside for evaluation. We then mix the training rows one to one with first messages taken from UltraChat, filtered to the maximum length in characters of the secret word data. Without this mix, the target model would no longer behave as a general assistant, which would make its activations unrepresentative.

Each adapter is a LoRA with rank $32$, $\alpha{=}64$, and dropout $0.05$, on every linear layer of the text part of the model. Gemma-3 and Qwen3.6 need an explicit pattern for these layers, for the reason given in \cref{app:qwen36-training}. A selection of all linear layers would also include the vision encoder and, on Qwen3.6, the routing projections of Gated DeltaNet, which get no gradient during training on text alone. We train with learning rate $5\times10^{-5}$ on a constant schedule with $5\%$ warmup, weight decay $0.01$, bfloat16, and a maximum sequence length of $1024$. Training lasts up to $10$ epochs, with early stopping after two evaluations without improvement of the validation loss. We keep the best checkpoint. One adapter takes about two hours on a single B200.

\section{Training the Qwen3.6-27B activation oracle}
\label{app:qwen36-training}

The Qwen3-8B oracle is the released checkpoint of \citet{karvonen_activation_2025}. For Qwen3.6-27B no oracle existed before this work. Qwen3.6-27B is a hybrid architecture, where $3/4$ of the layers that mix tokens use Gated DeltaNet and $1/4$ use Gated Attention. We adapt the LatentQA trainer of \citet{karvonen_activation_2025} to this architecture. We train the oracle with the same mix of data, which contains questions about system prompts, binary classification, and the prediction of the context, and with the same LoRA recipe ($r{=}8$, one epoch, bfloat16). The training needed two changes.

\paragraph{Layers of the LoRA.}
The original regular expression selects only \texttt{q\_proj}, \texttt{k\_proj}, \texttt{v\_proj}, \texttt{o\_proj}, and the projections of the MLP. These names exist only in the $1/4$ of the layers with full attention. The Gated DeltaNet layers use other names for their linear modules (\texttt{in\_proj\_qkv}, \texttt{in\_proj\_z}, \texttt{in\_proj\_b}, \texttt{in\_proj\_a}, \texttt{out\_proj}). With the original expression, LoRA covers only one quarter of the layers that mix tokens. The training loss then does not go down to the level of a standard model. Our expression covers both types of attention and still excludes the vision encoder and \texttt{lm\_head}.

\paragraph{Collection of the activations.}
The original code that collects activations from several layers took a window shifted by one position. We corrected the window to \texttt{[max\_offset:min\_offset]} everywhere. This correction is important when the activations come from the last tokens of long inputs.

With these two changes, the oracle gets a validation loss on data outside its training distribution comparable to the Qwen3-8B oracle (one B200, about $36$~hours, one epoch on about one million examples). We train the matching secret word target adapters with the recipe of \cref{app:taboo-training}.

\section{Hardware, software, and runtime}
\label{app:hwsw}

All runs use a single NVIDIA B200 GPU with bfloat16 weights. The full set of runs takes about $8$~hours on Qwen3-8B and about $48$~hours on Qwen3.6-27B. The two Gemma oracles and the later runs for forced choice, constrained label scoring, and the extra temperatures take about $70$~hours together. Software: PyTorch 2.11, Hugging Face Transformers 5.8, PEFT 0.19, flash-attn 2.8.3, Python 3.13.

\end{document}

%% file: scorecard-main-tbl.tex
\begin{table*}[t]
\centering
\footnotesize \setlength{\tabcolsep}{1.0pt}
\begin{tabular}{l rrr rrr rrr rrr}
\toprule
 & \multicolumn{3}{c}{Qwen3-8B} & \multicolumn{3}{c}{Qwen3.6-27B} & \multicolumn{3}{c}{Gemma-2-9B} & \multicolumn{3}{c}{Gemma-3-27B} \\
\cmidrule(lr){2-4}\cmidrule(lr){5-7}\cmidrule(lr){8-10}\cmidrule(lr){11-13}
Method & Acc & ECE & AUROC & Acc & ECE & AUROC & Acc & ECE & AUROC & Acc & ECE & AUROC \\
\midrule
Log-prob (offset) & .414 & .255\,\tiny$\pm$.005 & .840\,\tiny$\pm$.005 & .234 & .131\,\tiny$\pm$.005 & .835\,\tiny$\pm$.006 & .217 & \textbf{.037\,\tiny$\pm$.004} & .666\,\tiny$\pm$.008 & .433 & .151\,\tiny$\pm$.005 & .905\,\tiny$\pm$.004 \\
Bootstrap $T{=}1.0$ & .402 & \textbf{.057\,\tiny$\pm$.005} & .829\,\tiny$\pm$.005 & .225 & .147\,\tiny$\pm$.004 & .851\,\tiny$\pm$.007 & .219 & .177\,\tiny$\pm$.005 & .724\,\tiny$\pm$.008 & .429 & .134\,\tiny$\pm$.005 & .887\,\tiny$\pm$.004 \\
Bootstrap $T{=}1.5$ & .367 & .083\,\tiny$\pm$.005 & .824\,\tiny$\pm$.006 & .206 & .103\,\tiny$\pm$.004 & .812\,\tiny$\pm$.007 & .214 & .054\,\tiny$\pm$.004 & .748\,\tiny$\pm$.007 & .416 & .131\,\tiny$\pm$.005 & .864\,\tiny$\pm$.005 \\
Direct (numeric) & .414 & .582\,\tiny$\pm$.006 & .516\,\tiny$\pm$.003 & .234 & .753\,\tiny$\pm$.006 & .404\,\tiny$\pm$.007 & .217 & .783\,\tiny$\pm$.005 & .500\,\tiny$\pm$.000 & .433 & .562\,\tiny$\pm$.007 & .531\,\tiny$\pm$.003 \\
Constr.\ label $P(\mathrm{v.\,high})$ & .414 & .425\,\tiny$\pm$.006 & .812\,\tiny$\pm$.005 & .232 & \textbf{.069\,\tiny$\pm$.005} & .778\,\tiny$\pm$.006 & .217 & .181\,\tiny$\pm$.005 & .635\,\tiny$\pm$.009 & .433 & \textbf{.088\,\tiny$\pm$.005} & .813\,\tiny$\pm$.005 \\
Forced choice (M5) & .646 & .078\,\tiny$\pm$.004 & \textbf{.940\,\tiny$\pm$.003} & .533 & .080\,\tiny$\pm$.004 & \textbf{.936\,\tiny$\pm$.003} & .590 & .090\,\tiny$\pm$.004 & \textbf{.922\,\tiny$\pm$.003} & .550 & .131\,\tiny$\pm$.004 & \textbf{.959\,\tiny$\pm$.003} \\
\bottomrule
\end{tabular}
\caption{Accuracy, ECE, and AUROC of the five methods on all four oracles, with $6{,}000$ samples per row and oracle. ECE and AUROC come with one bootstrap standard error (1000 resamples). A lower ECE and a higher accuracy and AUROC are better. The best ECE and AUROC of each oracle are in \textbf{bold}. The three readouts built on the steering are in \cref{tab:scorecard-steered}. The full grid of methods is in \cref{app:full-scorecard}.}
\label{tab:scorecard}
\end{table*}

%% file: full-scorecard-qwen-tbl.tex
\begin{table*}[t]
\centering
\small \setlength{\tabcolsep}{4pt}
\begin{tabular}{lrrrrr|rrrrr}
\toprule
 & \multicolumn{5}{c|}{Qwen3-8B} & \multicolumn{5}{c}{Qwen3.6-27B} \\
Method & Acc & ECE & Brier & NLL & AUROC & Acc & ECE & Brier & NLL & AUROC \\
\midrule
Forced choice (M5) & \textbf{0.646} & 0.078 & \textbf{0.100} & \textbf{0.324} & \textbf{0.940} & \textbf{0.533} & 0.080 & \textbf{0.100} & \textbf{0.332} & \textbf{0.936} \\
Log-prob (no offset) & 0.414 & 0.256 & 0.249 & 0.748 & 0.824 & 0.234 & 0.131 & 0.157 & 0.484 & 0.842 \\
Log-prob (with offset) & 0.414 & 0.255 & 0.246 & 0.730 & 0.840 & 0.234 & 0.131 & 0.158 & 0.489 & 0.835 \\
Bootstrap $T{=}0.3$ & 0.419 & 0.334 & 0.303 & 2.636 & 0.784 & 0.233 & 0.479 & 0.369 & 2.367 & 0.837 \\
Bootstrap $T{=}0.5$ & 0.424 & 0.193 & 0.213 & 1.029 & 0.810 & 0.231 & 0.327 & 0.222 & 0.800 & 0.864 \\
Bootstrap $T{=}0.7$ & 0.414 & 0.097 & 0.171 & 0.558 & 0.830 & 0.230 & 0.218 & 0.157 & 0.514 & 0.863 \\
Bootstrap $T{=}1.0$ & 0.402 & \textbf{0.057} & 0.163 & 0.498 & 0.829 & 0.225 & 0.147 & 0.130 & 0.431 & 0.851 \\
Bootstrap $T{=}1.3$ & 0.387 & 0.076 & 0.171 & 0.519 & 0.823 & 0.209 & 0.125 & 0.125 & 0.415 & 0.837 \\
Bootstrap $T{=}1.5$ & 0.367 & 0.083 & 0.173 & 0.522 & 0.824 & 0.206 & 0.103 & 0.129 & 0.422 & 0.812 \\
Bootstrap $T{=}1.75$ & -- & -- & -- & -- & -- & 0.188 & 0.097 & 0.125 & 0.412 & 0.801 \\
Bootstrap $T{=}2.0$ & -- & -- & -- & -- & -- & 0.172 & 0.077 & 0.122 & 0.404 & 0.770 \\
Constrained label (expected) & 0.414 & 0.440 & 0.423 & 1.152 & 0.819 & 0.232 & 0.329 & 0.259 & 0.713 & 0.819 \\
Constrained label $P(\mathrm{high}{+})$ & 0.414 & 0.548 & 0.530 & 1.946 & 0.819 & 0.232 & 0.256 & 0.212 & 0.612 & 0.801 \\
Constrained label $P(\mathrm{very\ high})$ & 0.414 & 0.425 & 0.387 & 1.064 & 0.812 & 0.232 & \textbf{0.069} & 0.158 & 0.478 & 0.778 \\
MCMC agreement $T{=}0.125$ & 0.415 & 0.493 & 0.461 & 7.823 & 0.669 & 0.236 & 0.668 & 0.620 & 10.87 & 0.653 \\
MCMC agreement $T{=}0.25$ & 0.418 & 0.408 & 0.371 & 4.673 & 0.737 & 0.233 & 0.569 & 0.482 & 5.658 & 0.767 \\
MCMC agreement $T{=}0.5$ & 0.413 & 0.263 & 0.247 & 1.608 & 0.803 & 0.228 & 0.390 & 0.274 & 1.383 & 0.858 \\
Steering sensitivity & 0.418 & 0.404 & 0.354 & 4.401 & 0.763 & 0.232 & 0.501 & 0.413 & 4.172 & 0.725 \\
MCMC accept $T{=}0.125$ & 0.415 & 0.544 & 0.534 & 10.77 & 0.563 & 0.234 & 0.740 & 0.729 & 15.61 & 0.528 \\
MCMC accept $T{=}0.25$ & 0.403 & 0.547 & 0.532 & 10.28 & 0.579 & 0.230 & 0.738 & 0.722 & 15.04 & 0.547 \\
MCMC accept $T{=}0.5$ & 0.376 & 0.551 & 0.531 & 9.358 & 0.601 & 0.210 & 0.746 & 0.722 & 13.97 & 0.574 \\
Direct (numeric) & 0.414 & 0.582 & 0.580 & 12.66 & 0.516 & 0.234 & 0.753 & 0.752 & 16.02 & 0.404 \\
\bottomrule
\end{tabular}
\caption{All method rows on Qwen3-8B and Qwen3.6-27B, with $6{,}000$ samples per row. For ECE, Brier score, and NLL a lower value is better, while for accuracy and AUROC a higher value is better. The best value of each column is in \textbf{bold}. `--' marks a configuration that we did not run on this oracle, because we ran the extra temperatures only where the best ECE was at the edge of the original grid.}
\label{tab:full-scorecard}
\end{table*}

%% file: full-scorecard-gemma-tbl.tex
\begin{table*}[t]
\centering
\small \setlength{\tabcolsep}{4pt}
\begin{tabular}{lrrrrr|rrrrr}
\toprule
 & \multicolumn{5}{c|}{Gemma-2-9B} & \multicolumn{5}{c}{Gemma-3-27B} \\
Method & Acc & ECE & Brier & NLL & AUROC & Acc & ECE & Brier & NLL & AUROC \\
\midrule
Forced choice (M5) & \textbf{0.590} & 0.090 & \textbf{0.115} & \textbf{0.360} & \textbf{0.922} & \textbf{0.550} & 0.131 & \textbf{0.095} & \textbf{0.304} & \textbf{0.959} \\
Log-prob (no offset) & 0.217 & \textbf{0.037} & 0.162 & 0.501 & 0.666 & 0.433 & 0.151 & 0.161 & 0.487 & 0.905 \\
Log-prob (with offset) & 0.217 & \textbf{0.037} & 0.162 & 0.501 & 0.666 & 0.433 & 0.151 & 0.161 & 0.487 & 0.905 \\
Bootstrap $T{=}0.3$ & 0.218 & 0.613 & 0.559 & 6.403 & 0.617 & 0.434 & 0.453 & 0.412 & 5.459 & 0.743 \\
Bootstrap $T{=}0.5$ & 0.218 & 0.480 & 0.407 & 2.133 & 0.674 & 0.435 & 0.362 & 0.308 & 2.140 & 0.840 \\
Bootstrap $T{=}0.7$ & 0.222 & 0.338 & 0.283 & 0.944 & 0.689 & 0.435 & 0.260 & 0.218 & 0.821 & 0.880 \\
Bootstrap $T{=}1.0$ & 0.219 & 0.177 & 0.183 & 0.552 & 0.724 & 0.429 & 0.134 & 0.153 & 0.473 & 0.887 \\
Bootstrap $T{=}1.3$ & 0.219 & 0.079 & 0.156 & 0.486 & 0.722 & 0.422 & 0.117 & 0.156 & 0.485 & 0.876 \\
Bootstrap $T{=}1.5$ & 0.214 & 0.054 & 0.145 & 0.459 & 0.748 & 0.416 & 0.131 & 0.173 & 0.522 & 0.864 \\
Bootstrap $T{=}1.75$ & 0.202 & 0.040 & 0.142 & 0.450 & 0.743 & -- & -- & -- & -- & -- \\
Bootstrap $T{=}2.0$ & 0.204 & 0.050 & 0.145 & 0.455 & 0.738 & -- & -- & -- & -- & -- \\
Constrained label (expected) & 0.217 & 0.463 & 0.374 & 0.958 & 0.671 & 0.433 & 0.318 & 0.312 & 0.833 & 0.821 \\
Constrained label $P(\mathrm{high}{+})$ & 0.217 & 0.391 & 0.311 & 0.827 & 0.680 & 0.433 & 0.416 & 0.373 & 1.121 & 0.815 \\
Constrained label $P(\mathrm{very\ high})$ & 0.217 & 0.181 & 0.198 & 0.584 & 0.635 & 0.433 & \textbf{0.088} & 0.183 & 0.548 & 0.813 \\
MCMC agreement $T{=}0.125$ & 0.217 & 0.722 & 0.700 & 13.66 & 0.546 & 0.436 & 0.527 & 0.510 & 10.27 & 0.592 \\
MCMC agreement $T{=}0.25$ & 0.217 & 0.657 & 0.616 & 9.985 & 0.581 & 0.437 & 0.479 & 0.447 & 7.466 & 0.679 \\
MCMC agreement $T{=}0.5$ & 0.214 & 0.504 & 0.435 & 3.767 & 0.652 & 0.435 & 0.374 & 0.324 & 3.049 & 0.808 \\
Steering sensitivity & 0.208 & 0.649 & 0.618 & 10.86 & 0.495 & 0.433 & 0.469 & 0.433 & 7.421 & 0.682 \\
MCMC accept $T{=}0.125$ & 0.217 & 0.714 & 0.689 & 13.27 & 0.553 & 0.436 & 0.516 & 0.516 & 7.712 & 0.487 \\
MCMC accept $T{=}0.25$ & 0.209 & 0.743 & 0.722 & 14.30 & 0.542 & 0.437 & 0.520 & 0.517 & 7.784 & 0.518 \\
MCMC accept $T{=}0.5$ & 0.194 & 0.770 & 0.753 & 15.19 & 0.533 & 0.422 & 0.532 & 0.527 & 7.159 & 0.526 \\
Direct (numeric) & 0.217 & 0.783 & 0.783 & 18.02 & 0.500 & 0.433 & 0.562 & 0.559 & 12.10 & 0.531 \\
\bottomrule
\end{tabular}
\caption{All method rows on Gemma-2-9B and Gemma-3-27B, with $6{,}000$ samples per row. For ECE, Brier score, and NLL a lower value is better, while for accuracy and AUROC a higher value is better. The best value of each column is in \textbf{bold}. `--' marks a configuration that we did not run on this oracle, because we ran the extra temperatures only where the best ECE was at the edge of the original grid.}
\label{tab:full-scorecard-gemma}
\end{table*}

%% file: bootstrap-cis-tbl.tex
\begin{table*}[t]
\centering
\small \setlength{\tabcolsep}{4pt}
\begin{tabular}{lrrrr}
\toprule
Method & ECE & Brier & NLL & AUROC \\
\midrule
\multicolumn{5}{l}{\textit{Qwen3-8B}} \\
Forced choice (M5) & .078\,\scriptsize[.070,.086] & .100\,\scriptsize[.096,.104] & .324\,\scriptsize[.314,.333] & .940\,\scriptsize[.934,.946] \\
Log-prob (with offset) & .255\,\scriptsize[.244,.266] & .246\,\scriptsize[.238,.254] & .730\,\scriptsize[.707,.754] & .840\,\scriptsize[.831,.850] \\
Bootstrap $T{=}1.0$ & .057\,\scriptsize[.048,.067] & .163\,\scriptsize[.159,.168] & .498\,\scriptsize[.488,.509] & .829\,\scriptsize[.819,.839] \\
Bootstrap $T{=}1.5$ & .083\,\scriptsize[.076,.093] & .173\,\scriptsize[.168,.178] & .522\,\scriptsize[.510,.533] & .824\,\scriptsize[.813,.836] \\
Constrained label $P(\mathrm{very\ high})$ & .425\,\scriptsize[.413,.437] & .387\,\scriptsize[.378,.395] & 1.064\,\scriptsize[1.040,1.087] & .812\,\scriptsize[.802,.823] \\
Direct (numeric) & .582\,\scriptsize[.571,.595] & .580\,\scriptsize[.569,.593] & 12.658\,\scriptsize[12.397,12.938] & .516\,\scriptsize[.510,.521] \\
\midrule
\multicolumn{5}{l}{\textit{Qwen3.6-27B}} \\
Forced choice (M5) & .080\,\scriptsize[.073,.088] & .100\,\scriptsize[.096,.104] & .332\,\scriptsize[.323,.342] & .936\,\scriptsize[.930,.943] \\
Log-prob (with offset) & .131\,\scriptsize[.122,.141] & .158\,\scriptsize[.151,.166] & .489\,\scriptsize[.469,.511] & .835\,\scriptsize[.822,.847] \\
Bootstrap $T{=}1.0$ & .147\,\scriptsize[.139,.156] & .130\,\scriptsize[.127,.134] & .431\,\scriptsize[.423,.439] & .851\,\scriptsize[.839,.864] \\
Bootstrap $T{=}1.5$ & .103\,\scriptsize[.095,.112] & .129\,\scriptsize[.125,.133] & .422\,\scriptsize[.412,.432] & .812\,\scriptsize[.798,.826] \\
Constrained label $P(\mathrm{very\ high})$ & .069\,\scriptsize[.061,.079] & .158\,\scriptsize[.153,.165] & .478\,\scriptsize[.464,.493] & .778\,\scriptsize[.764,.790] \\
Direct (numeric) & .753\,\scriptsize[.741,.764] & .752\,\scriptsize[.742,.763] & 16.022\,\scriptsize[15.756,16.293] & .404\,\scriptsize[.391,.416] \\
\midrule
\multicolumn{5}{l}{\textit{Gemma-2-9B}} \\
Forced choice (M5) & .090\,\scriptsize[.082,.098] & .115\,\scriptsize[.111,.118] & .360\,\scriptsize[.352,.369] & .922\,\scriptsize[.915,.929] \\
Log-prob (with offset) & .037\,\scriptsize[.031,.047] & .162\,\scriptsize[.156,.168] & .501\,\scriptsize[.485,.517] & .666\,\scriptsize[.650,.682] \\
Bootstrap $T{=}1.0$ & .177\,\scriptsize[.167,.186] & .183\,\scriptsize[.179,.187] & .552\,\scriptsize[.541,.564] & .724\,\scriptsize[.708,.739] \\
Bootstrap $T{=}1.5$ & .054\,\scriptsize[.048,.064] & .145\,\scriptsize[.141,.150] & .459\,\scriptsize[.449,.470] & .748\,\scriptsize[.733,.761] \\
Constrained label $P(\mathrm{very\ high})$ & .181\,\scriptsize[.173,.193] & .198\,\scriptsize[.194,.202] & .584\,\scriptsize[.575,.593] & .635\,\scriptsize[.616,.653] \\
Direct (numeric) & .783\,\scriptsize[.772,.794] & .783\,\scriptsize[.772,.794] & 18.022\,\scriptsize[17.780,18.275] & .500\,\scriptsize[.500,.500] \\
\midrule
\multicolumn{5}{l}{\textit{Gemma-3-27B}} \\
Forced choice (M5) & .131\,\scriptsize[.124,.138] & .095\,\scriptsize[.092,.099] & .304\,\scriptsize[.295,.313] & .959\,\scriptsize[.954,.964] \\
Log-prob (with offset) & .151\,\scriptsize[.142,.160] & .161\,\scriptsize[.157,.165] & .487\,\scriptsize[.477,.497] & .905\,\scriptsize[.897,.912] \\
Bootstrap $T{=}1.0$ & .134\,\scriptsize[.124,.144] & .153\,\scriptsize[.150,.157] & .473\,\scriptsize[.464,.482] & .887\,\scriptsize[.879,.896] \\
Bootstrap $T{=}1.5$ & .131\,\scriptsize[.121,.141] & .173\,\scriptsize[.169,.177] & .522\,\scriptsize[.512,.531] & .864\,\scriptsize[.854,.873] \\
Constrained label $P(\mathrm{very\ high})$ & .088\,\scriptsize[.078,.099] & .183\,\scriptsize[.180,.187] & .548\,\scriptsize[.541,.556] & .813\,\scriptsize[.802,.823] \\
Direct (numeric) & .562\,\scriptsize[.548,.575] & .559\,\scriptsize[.545,.572] & 12.102\,\scriptsize[11.794,12.394] & .531\,\scriptsize[.526,.536] \\
\bottomrule
\end{tabular}
\caption{Bootstrap $95\%$ confidence intervals (1000 resamples) for the main method rows on all four oracles. Each cell gives the value, followed by the 2.5 and 97.5 percentiles. Each row has $6{,}000$ samples.}
\label{tab:bootstrap-cis}
\end{table*}

%% file: resampling-tbl.tex
\begin{table*}[t]
\centering
\small \setlength{\tabcolsep}{4pt}
\begin{tabular}{ll rrrr rrrr}
\toprule
 & & \multicolumn{4}{c}{Resample secret words} & \multicolumn{4}{c}{Resample words, contexts, questions} \\
\cmidrule(lr){3-6}\cmidrule(lr){7-10}
Metric & Ordering & Q-8B & Q-27B & G-9B & G-27B & Q-8B & Q-27B & G-9B & G-27B \\
\midrule
AUROC & Forced choice $>$ log probability & 100.0 & 100.0 & 100.0 & 100.0 & 100.0 & 98.1 & 100.0 & 99.4 \\
AUROC & Forced choice $>$ bootstrap & 100.0 & 100.0 & 100.0 & 100.0 & 100.0 & 100.0 & 100.0 & 100.0 \\
AUROC & Log probability $>$ direct & 100.0 & 100.0 & 100.0 & 100.0 & 100.0 & 100.0 & 99.3 & 100.0 \\
AUROC & Bootstrap $>$ direct & 100.0 & 100.0 & 100.0 & 100.0 & 100.0 & 100.0 & 100.0 & 100.0 \\
AUROC & Constrained label $>$ direct & 100.0 & 100.0 & 99.8 & 100.0 & 100.0 & 100.0 & 98.0 & 100.0 \\
ECE & Bootstrap $<$ direct & 100.0 & 100.0 & 100.0 & 100.0 & 100.0 & 100.0 & 100.0 & 100.0 \\
ECE & Forced choice $<$ direct & 100.0 & 100.0 & 100.0 & 100.0 & 100.0 & 100.0 & 100.0 & 100.0 \\
\bottomrule
\end{tabular}
\caption{Share of 1000 resamples (\%) in which each comparison gives the same result. Left: we resample whole secret words, so that the 300 samples of a word are kept together. Right: we resample secret words, context prompts, and verbalizer prompts together \citep{owen_pigeonhole_2007}. The bootstrap is at the tuned temperature of each oracle. Constrained label scoring uses the $P(\mathrm{very\ high})$ readout. Q-8B: Qwen3-8B, Q-27B: Qwen3.6-27B, G-9B: Gemma-2-9B, G-27B: Gemma-3-27B.}
\label{tab:resampling}
\end{table*}

%% file: errors-tbl.tex
\begin{table}[t]
\centering
\small \setlength{\tabcolsep}{3pt}
\begin{tabularx}{\linewidth}{X rrrr}
\toprule
 & \rotatebox{70}{Qwen3-8B} & \rotatebox{70}{Qwen3.6-27B} & \rotatebox{70}{Gemma-2-9B} & \rotatebox{70}{Gemma-3-27B} \\
\midrule
Incorrect answers in free text that name none of the 20 words & 98 & 100 & 100 & 100 \\
\quad among these, forced choice finds the secret word & 40 & 39 & 48 & 32 \\
Samples where no bootstrap answer names the secret word & 44 & 64 & 56 & 46 \\
\quad among these, forced choice finds the secret word & 23 & 28 & 30 & 24 \\
\bottomrule
\end{tabularx}
\caption{Where the forced choice gain comes from (\%). Top: greedy answers in free text (M1). Bottom: the twenty bootstrap answers at $T{=}1.0$ (M2).}
\label{tab:errors}
\end{table}

%% file: failures-tbl.tex
\begin{table*}[t]
\centering
\small \setlength{\tabcolsep}{4pt}
\begin{tabular}{l rrr rrr rrr rrr}
\toprule
 & \multicolumn{3}{c}{Qwen3-8B} & \multicolumn{3}{c}{Qwen3.6-27B} & \multicolumn{3}{c}{Gemma-2-9B} & \multicolumn{3}{c}{Gemma-3-27B} \\
\cmidrule(lr){2-4}\cmidrule(lr){5-7}\cmidrule(lr){8-10}\cmidrule(lr){11-13}
Incorrect answer & \% & LP & BS & \% & LP & BS & \% & LP & BS & \% & LP & BS \\
\midrule
Copies the context prompt & 20 & .791 & .745 & 15 & .874 & .790 & 32 & .663 & .685 & 27 & .904 & .858 \\
Copies the oracle question & 22 & .886 & .849 & 30 & .837 & .853 & 20 & .908 & .841 & 4 & .974 & .950 \\
Other & 59 & .840 & .843 & 55 & .822 & .869 & 47 & .563 & .702 & 69 & .901 & .894 \\
\bottomrule
\end{tabular}
\caption{The failure types of \citet{jakkli_current_2026} inside the incorrect answers in free text. \%: share of the incorrect greedy answers of log probability (M1) in each type. LP and BS: AUROC of log probability and of the bootstrap at $T{=}1.0$ (M2) when the correct answers are compared only with the incorrect answers of this type. An answer copies the context prompt when its word is in the context prompt but not in the question of the oracle.}
\label{tab:failures}
\end{table*}

%% file: postcal-qwen3-8b-tbl.tex
\begin{table*}[t]
\centering
\small \setlength{\tabcolsep}{4pt}
\begin{tabular}{lrrrrr|rrrrr}
\toprule
 & \multicolumn{5}{c|}{Split by word} & \multicolumn{5}{c}{Random split} \\
Method & None & Temp & Platt & Iso & Beta & None & Temp & Platt & Iso & Beta \\
\midrule
Forced choice (M5) & .076 & .041 & .040 & .043 & .041 & .081 & .037 & .016 & .012 & .014 \\
Log-prob (with offset) & .172 & .180 & .128 & .127 & .128 & .258 & .157 & .018 & .021 & .017 \\
Bootstrap $T{=}0.7$ & .156 & .159 & .118 & .113 & .111 & .092 & .089 & .029 & .025 & .038 \\
Bootstrap $T{=}1.0$ & .061 & .033 & .094 & .094 & .095 & .067 & .031 & .024 & .029 & .024 \\
Bootstrap $T{=}1.5$ & .069 & .072 & .078 & .077 & .077 & .087 & .093 & .020 & .019 & .015 \\
Constrained label $P(\mathrm{very\ high})$ & .525 & .295 & .152 & .149 & .151 & .423 & .107 & .026 & .016 & .016 \\
Direct (numeric) & .691 & .354 & .219 & .217 & .217 & .579 & .241 & .009 & .007 & .007 \\
MCMC accept $T{=}0.125$ & .642 & .327 & .197 & .197 & .197 & .541 & .223 & .007 & .012 & .010 \\
MCMC agreement $T{=}0.5$ & .334 & .274 & .145 & .135 & .133 & .261 & .176 & .060 & .023 & .024 \\
Steering sensitivity & .485 & .307 & .136 & .135 & .135 & .399 & .172 & .013 & .013 & .016 \\
\bottomrule
\end{tabular}
\caption{Post hoc calibration on Qwen3-8B: ECE on the test data after each calibrator is fit on separate data. \emph{Split by word}: fit on 10 of the 20 secret words and test on the other 10. \emph{Random split}: random halves of the samples, the same for every method. \emph{None}: no calibrator. A lower value is better.}
\label{tab:postcal-qwen3-8b}
\end{table*}

%% file: postcal-qwen3-6-27b-tbl.tex
\begin{table*}[t]
\centering
\small \setlength{\tabcolsep}{4pt}
\begin{tabular}{lrrrrr|rrrrr}
\toprule
 & \multicolumn{5}{c|}{Split by word} & \multicolumn{5}{c}{Random split} \\
Method & None & Temp & Platt & Iso & Beta & None & Temp & Platt & Iso & Beta \\
\midrule
Forced choice (M5) & .088 & .055 & .036 & .029 & .025 & .090 & .054 & .024 & .027 & .022 \\
Log-prob (with offset) & .142 & .090 & .030 & .017 & .021 & .138 & .096 & .030 & .019 & .021 \\
Bootstrap $T{=}0.7$ & .209 & .156 & .026 & .016 & .030 & .210 & .160 & .024 & .017 & .034 \\
Bootstrap $T{=}1.0$ & .145 & .017 & .012 & .019 & .012 & .152 & .024 & .027 & .022 & .026 \\
Bootstrap $T{=}1.5$ & .101 & .054 & .016 & .022 & .020 & .103 & .042 & .014 & .010 & .007 \\
Constrained label $P(\mathrm{very\ high})$ & .084 & .077 & .057 & .030 & .027 & .073 & .070 & .041 & .017 & .031 \\
Direct (numeric) & .740 & .399 & .035 & .028 & .042 & .744 & .403 & .017 & .017 & .016 \\
MCMC accept $T{=}0.125$ & .728 & .404 & .023 & .023 & .023 & .733 & .409 & .014 & .014 & .015 \\
MCMC agreement $T{=}0.5$ & .381 & .319 & .044 & .022 & .022 & .380 & .319 & .041 & .016 & .016 \\
Steering sensitivity & .491 & .306 & .027 & .025 & .025 & .494 & .315 & .017 & .018 & .018 \\
\bottomrule
\end{tabular}
\caption{Post hoc calibration on Qwen3.6-27B: ECE on the test data after each calibrator is fit on separate data. \emph{Split by word}: fit on 10 of the 20 secret words and test on the other 10. \emph{Random split}: random halves of the samples, the same for every method. \emph{None}: no calibrator. A lower value is better.}
\label{tab:postcal-qwen3-6-27b}
\end{table*}

%% file: postcal-gemma-2-9b-tbl.tex
\begin{table*}[t]
\centering
\small \setlength{\tabcolsep}{4pt}
\begin{tabular}{lrrrrr|rrrrr}
\toprule
 & \multicolumn{5}{c|}{Split by word} & \multicolumn{5}{c}{Random split} \\
Method & None & Temp & Platt & Iso & Beta & None & Temp & Platt & Iso & Beta \\
\midrule
Forced choice (M5) & .122 & .072 & .097 & .093 & .092 & .092 & .031 & .019 & .015 & .014 \\
Log-prob (with offset) & .072 & .094 & .101 & .100 & .102 & .046 & .040 & .026 & .021 & .027 \\
Bootstrap $T{=}0.7$ & .391 & .362 & .121 & .110 & .115 & .344 & .294 & .033 & .027 & .032 \\
Bootstrap $T{=}1.0$ & .225 & .199 & .099 & .099 & .099 & .178 & .148 & .010 & .021 & .021 \\
Bootstrap $T{=}1.5$ & .096 & .074 & .095 & .096 & .095 & .056 & .018 & .014 & .013 & .014 \\
Constrained label $P(\mathrm{very\ high})$ & .243 & .225 & .128 & .121 & .118 & .185 & .172 & .030 & .011 & .017 \\
Direct (numeric) & .852 & .518 & .139 & .139 & .139 & .789 & .455 & .013 & .013 & .013 \\
MCMC accept $T{=}0.125$ & .780 & .477 & .138 & .137 & .137 & .722 & .416 & .013 & .013 & .013 \\
MCMC agreement $T{=}0.5$ & .564 & .396 & .130 & .128 & .128 & .513 & .335 & .014 & .015 & .015 \\
Steering sensitivity & .716 & .458 & .138 & .138 & .137 & .658 & .399 & .018 & .017 & .017 \\
\bottomrule
\end{tabular}
\caption{Post hoc calibration on Gemma-2-9B: ECE on the test data after each calibrator is fit on separate data. \emph{Split by word}: fit on 10 of the 20 secret words and test on the other 10. \emph{Random split}: random halves of the samples, the same for every method. \emph{None}: no calibrator. A lower value is better.}
\label{tab:postcal-gemma-2-9b}
\end{table*}

%% file: postcal-gemma-3-27b-tbl.tex
\begin{table*}[t]
\centering
\small \setlength{\tabcolsep}{4pt}
\begin{tabular}{lrrrrr|rrrrr}
\toprule
 & \multicolumn{5}{c|}{Split by word} & \multicolumn{5}{c}{Random split} \\
Method & None & Temp & Platt & Iso & Beta & None & Temp & Platt & Iso & Beta \\
\midrule
Forced choice (M5) & .128 & .090 & .016 & .014 & .013 & .133 & .093 & .012 & .015 & .017 \\
Log-prob (with offset) & .184 & .170 & .040 & .038 & .038 & .152 & .146 & .021 & .024 & .018 \\
Bootstrap $T{=}0.7$ & .231 & .193 & .098 & .029 & .046 & .262 & .222 & .094 & .028 & .043 \\
Bootstrap $T{=}1.0$ & .123 & .087 & .035 & .038 & .035 & .136 & .106 & .020 & .024 & .023 \\
Bootstrap $T{=}1.5$ & .162 & .152 & .047 & .047 & .047 & .128 & .121 & .015 & .016 & .014 \\
Constrained label $P(\mathrm{very\ high})$ & .081 & .045 & .085 & .091 & .087 & .094 & .045 & .024 & .021 & .023 \\
Direct (numeric) & .512 & .176 & .110 & .109 & .110 & .566 & .230 & .009 & .009 & .009 \\
MCMC accept $T{=}0.125$ & .464 & .113 & .103 & .103 & .103 & .519 & .169 & .009 & .009 & .009 \\
MCMC agreement $T{=}0.5$ & .338 & .208 & .050 & .054 & .055 & .378 & .189 & .034 & .009 & .013 \\
Steering sensitivity & .425 & .138 & .089 & .090 & .090 & .471 & .189 & .003 & .003 & .003 \\
\bottomrule
\end{tabular}
\caption{Post hoc calibration on Gemma-3-27B: ECE on the test data after each calibrator is fit on separate data. \emph{Split by word}: fit on 10 of the 20 secret words and test on the other 10. \emph{Random split}: random halves of the samples, the same for every method. \emph{None}: no calibrator. A lower value is better.}
\label{tab:postcal-gemma-3-27b}
\end{table*}

%% file: tsweep-tbl.tex
\begin{table*}[t]
\centering
\small \setlength{\tabcolsep}{4pt}
\begin{tabular}{l rr rr rr rr}
\toprule
 & \multicolumn{2}{c}{Qwen3-8B} & \multicolumn{2}{c}{Qwen3.6-27B} & \multicolumn{2}{c}{Gemma-2-9B} & \multicolumn{2}{c}{Gemma-3-27B} \\
\cmidrule(lr){2-3}\cmidrule(lr){4-5}\cmidrule(lr){6-7}\cmidrule(lr){8-9}
$T$ & Acc & ECE & Acc & ECE & Acc & ECE & Acc & ECE \\
\midrule
0.3 & .419 & .334 & .233 & .479 & .218 & .613 & .434 & .453 \\
0.5 & .424 & .193 & .231 & .327 & .218 & .480 & .435 & .362 \\
0.7 & .414 & .097 & .230 & .218 & .222 & .338 & .435 & .260 \\
1.0 & .402 & \textbf{.057} & .225 & .147 & .219 & .177 & .429 & .134 \\
1.3 & .387 & .076 & .209 & .125 & .219 & .079 & .422 & \textbf{.117} \\
1.5 & .367 & .083 & .206 & .103 & .214 & .054 & .416 & .131 \\
1.75 & -- & -- & .188 & .097 & .202 & \textbf{.040} & -- & -- \\
2.0 & -- & -- & .172 & \textbf{.077} & .204 & .050 & -- & -- \\
\bottomrule
\end{tabular}
\caption{The bootstrap at each temperature, with $k{=}20$ answers for each sample. The best ECE of each oracle is in \textbf{bold}. We ran the temperatures above $1.5$ on the two oracles whose ECE was still going down at $T{=}1.5$.}
\label{tab:tsweep}
\end{table*}

%% file: tsel-tbl.tex
\begin{table}[t]
\centering
\small \setlength{\tabcolsep}{4pt}
\begin{tabular}{l rrrrr}
\toprule
 & No labels & \multicolumn{3}{c}{Labeled words} & Best \\
\cmidrule(lr){3-5}
Oracle & $T{=}1.0$ & 1 & 5 & 10 & $T$ \\
\midrule
Qwen3-8B & .061 & .109 & .075 & .074 & .057 \\
Qwen3.6-27B & .148 & .095 & .088 & .081 & .077 \\
Gemma-2-9B & .176 & .110 & .069 & .072 & .040 \\
Gemma-3-27B & .135 & .127 & .124 & .122 & .117 \\
\bottomrule
\end{tabular}
\caption{ECE of the bootstrap when the temperature is chosen with the rule of \cref{sec:results-temperature} on 1, 5, or 10 labeled secret words, computed on the other words and averaged over 200 random choices of the labeled words. \emph{No labels}: the default $T{=}1.0$. \emph{Best $T$}: the temperature of the grid with the lowest ECE on all 20 words.}
\label{tab:tsel}
\end{table}

%% file: k-ablation-tbl.tex
\begin{table*}[t]
\centering
\small \setlength{\tabcolsep}{4pt}
\begin{tabular}{l rrr rrr rrr rrr}
\toprule
 & \multicolumn{3}{c}{Qwen3-8B ($T{=}1.0$)} & \multicolumn{3}{c}{Qwen3.6-27B ($T{=}1.5$)} & \multicolumn{3}{c}{Gemma-2-9B ($T{=}1.5$)} & \multicolumn{3}{c}{Gemma-3-27B ($T{=}1.3$)} \\
\cmidrule(lr){2-4}\cmidrule(lr){5-7}\cmidrule(lr){8-10}\cmidrule(lr){11-13}
$k$ & Acc & ECE & AUROC & Acc & ECE & AUROC & Acc & ECE & AUROC & Acc & ECE & AUROC \\
\midrule
3 & .285 & .240 & .808 & .107 & .313 & .735 & .112 & .307 & .698 & .298 & .223 & .814 \\
5 & .334 & .131 & .823 & .138 & .204 & .766 & .141 & .198 & .745 & .357 & .110 & .824 \\
10 & .371 & .069 & .835 & .173 & .131 & .797 & .184 & .101 & .739 & .404 & .097 & .853 \\
20 & .402 & .057 & .829 & .206 & .103 & .812 & .214 & .054 & .748 & .422 & .117 & .876 \\
\bottomrule
\end{tabular}
\caption{The bootstrap confidence computed from the first $k$ of the 20 saved answers, at the temperature with the best ECE in the original grid of each oracle. Most of the gain in calibration and AUROC is already present at $k{=}10$. The number $k$ is the main factor in the cost of the bootstrap.}
\label{tab:kablation}
\end{table*}

%% file: perword-tbl.tex
\begin{table}[t]
\centering
\scriptsize \setlength{\tabcolsep}{2pt}
\begin{tabular}{lrrrr}
\toprule
Word & Qwen3-8B & Qwen3.6-27B & Gemma-2-9B & Gemma-3-27B \\
\midrule
moon & .827 & .380 & .623 & .467 \\
snow & .727 & .413 & .320 & .487 \\
jump & .627 & .240 & .633 & .553 \\
smile & .610 & .240 & .623 & .423 \\
ship & .580 & .100 & .263 & .473 \\
chair & .557 & .190 & .133 & .513 \\
flag & .523 & .097 & .513 & .477 \\
salt & .477 & .277 & .060 & .527 \\
green & .473 & .450 & .213 & .497 \\
dance & .350 & .197 & .043 & .470 \\
book & .337 & .267 & .153 & .323 \\
gold & .327 & .247 & .047 & .300 \\
cloud & .307 & .163 & .073 & .593 \\
flame & .297 & .147 & .070 & .170 \\
leaf & .240 & .050 & .087 & .373 \\
blue & .180 & .500 & .157 & .560 \\
wave & .170 & .110 & .000 & .467 \\
clock & .170 & .157 & .277 & .277 \\
song & .150 & .100 & .000 & .047 \\
rock & .120 & .170 & .087 & .587 \\
\bottomrule
\end{tabular}
\caption{Accuracy of each secret word with the bootstrap at $T{=}1.0$ on all four oracles, sorted by the accuracy on Qwen3-8B. On every oracle, the accuracy changes by an order of magnitude between words.}
\label{tab:perword-bothmodels}
\end{table}

%% file: scorecard-steered-tbl.tex
\begin{table*}[t]
\centering
\footnotesize \setlength{\tabcolsep}{1.0pt}
\begin{tabular}{l rrr rrr rrr rrr}
\toprule
 & \multicolumn{3}{c}{Qwen3-8B} & \multicolumn{3}{c}{Qwen3.6-27B} & \multicolumn{3}{c}{Gemma-2-9B} & \multicolumn{3}{c}{Gemma-3-27B} \\
\cmidrule(lr){2-4}\cmidrule(lr){5-7}\cmidrule(lr){8-10}\cmidrule(lr){11-13}
Method & Acc & ECE & AUROC & Acc & ECE & AUROC & Acc & ECE & AUROC & Acc & ECE & AUROC \\
\midrule
MCMC agreement $T{=}0.5$ & .413 & \textbf{.263\,\tiny$\pm$.005} & \textbf{.803\,\tiny$\pm$.006} & .228 & \textbf{.390\,\tiny$\pm$.005} & \textbf{.858\,\tiny$\pm$.006} & .214 & \textbf{.504\,\tiny$\pm$.006} & \textbf{.652\,\tiny$\pm$.008} & .435 & \textbf{.374\,\tiny$\pm$.005} & \textbf{.808\,\tiny$\pm$.005} \\
Steering sensitivity & .418 & .404\,\tiny$\pm$.006 & .763\,\tiny$\pm$.005 & .232 & .501\,\tiny$\pm$.005 & .725\,\tiny$\pm$.007 & .208 & .649\,\tiny$\pm$.006 & .495\,\tiny$\pm$.008 & .433 & .469\,\tiny$\pm$.006 & .682\,\tiny$\pm$.005 \\
MCMC accept $T{=}0.125$ & .415 & .544\,\tiny$\pm$.006 & .563\,\tiny$\pm$.005 & .234 & .740\,\tiny$\pm$.006 & .528\,\tiny$\pm$.004 & .217 & .714\,\tiny$\pm$.005 & .553\,\tiny$\pm$.006 & .436 & .516\,\tiny$\pm$.007 & .487\,\tiny$\pm$.007 \\
\bottomrule
\end{tabular}
\caption{The three confidence readouts built on the steering, with $6{,}000$ samples per row and oracle, each at the best temperature of its grid. ECE and AUROC come with one bootstrap standard error (1000 resamples). The best ECE and AUROC of each oracle among these three readouts are in \textbf{bold}. Every row here is less useful than the tuned bootstrap of \cref{tab:scorecard}, both in ECE and in AUROC.}
\label{tab:scorecard-steered}
\end{table*}

%% file: custom.bib
@misc{karvonen_activation_2025,
  title      = {Activation oracles: training and evaluating {LLMs} as general-purpose activation explainers},
  shorttitle = {Activation oracles},
  author     = {Karvonen, Adam and Chua, James and Dumas, Clément and Fraser-Taliente, Kit and Kantamneni, Subhash and Minder, Julian and Ong, Euan and Sharma, Arnab Sen and Wen, Daniel and Evans, Owain and Marks, Samuel},
  year       = 2025,
  publisher  = {arXiv},
  doi        = {10.48550/ARXIV.2512.15674},
  url        = {https://arxiv.org/abs/2512.15674},
  urldate    = {2026-03-20},
  copyright  = {Creative Commons Attribution 4.0 International},
  language   = {en}
}

@article{bricken_building_2025,
  title     = {Building and evaluating alignment auditing agents},
  author    = {Bricken, Trenton and Wang, Rowan and Bowman, Sam and Ong, Euan and Treutlein, Johannes and Wu, Jeff and Hubinger, Evan and Marks, Samuel},
  year      = 2025,
  journal   = {Alignment Science Blog},
  publisher = {Anthropic},
  url       = {https://alignment.anthropic.com/2025/automated-auditing/},
  language  = {en}
}

@inproceedings{karan_reasoning_2025,
  title     = {Reasoning with Sampling: Your Base Model is Smarter Than You Think},
  author    = {Aayush Karan and Yilun Du},
  year      = 2026,
  booktitle = {The Fourteenth International Conference on Learning Representations},
  url       = {https://openreview.net/forum?id=Vsgq2ldr4K}
}

@article{naeini_obtaining_2015,
  title    = {Obtaining well calibrated probabilities using bayesian binning},
  author   = {Naeini, Mahdi Pakdaman and Cooper, Gregory F. and Hauskrecht, Milos},
  year     = 2015,
  month    = feb,
  journal  = {Proceedings of the 29th AAAI Conference on Artificial Intelligence},
  volume   = 29,
  number   = 1,
  pages    = {1--7},
  doi      = {10.1609/aaai.v29i1.9602},
  issn     = {2374-3468, 2159-5399},
  url      = {https://ojs.aaai.org/index.php/AAAI/article/view/9602},
  language = {en}
}

@article{brier_verification_1950,
  title    = {Verification of forecasts expressed in terms of probability},
  author   = {Brier, Glenn W.},
  year     = 1950,
  journal  = {Monthly Weather Review},
  volume   = 78,
  number   = 1,
  pages    = {1--3},
  doi      = {10.1175/1520-0493(1950)078<0001:VOFEIT>2.0.CO;2},
  issn     = {0027-0644, 1520-0493},
  url      = {http://journals.ametsoc.org/doi/10.1175/1520-0493(1950)078<0001:VOFEIT>2.0.CO;2},
  language = {en}
}

@misc{yuan_hidden_2026,
  title      = {Hidden error awareness in chain-of-thought reasoning: the signal is diagnostic, not causal},
  shorttitle = {Hidden error awareness},
  author     = {Yuan, Aojie and Su, Zhiyuan Julian and Zhang, Haiyue and Nian, Yi and Zhao, Yue},
  year       = 2026,
  doi        = {10.48550/arXiv.2605.09502},
  url        = {https://arxiv.org/abs/2605.09502},
  language   = {en}
}

@misc{miao_closing_2026,
  title    = {Closing the confidence-faithfulness gap in large language models},
  author   = {Miao, Miranda Muqing and Ungar, Lyle},
  year     = 2026,
  doi      = {10.48550/arXiv.2603.25052},
  url      = {https://arxiv.org/abs/2603.25052},
  language = {en}
}

@misc{turner_steering_2024,
  title    = {Steering language models with activation engineering},
  author   = {Turner, Alexander Matt and Thiergart, Lisa and Leech, Gavin and Udell, David and Vazquez, Juan J. and Mini, Ulisse and MacDiarmid, Monte},
  year     = 2024,
  doi      = {10.48550/arXiv.2308.10248},
  url      = {https://arxiv.org/abs/2308.10248},
  language = {en}
}

@misc{sheshadri_auditbench_2026,
  title      = {{AuditBench}: evaluating alignment auditing techniques on models with hidden behaviors},
  shorttitle = {{AuditBench}},
  author     = {Sheshadri, Abhay and Ewart, Aidan and Fronsdal, Kai and Gupta, Isha and Bowman, Samuel R. and Price, Sara and Marks, Samuel and Wang, Rowan},
  year       = 2026,
  doi        = {10.48550/arXiv.2602.22755},
  url        = {https://arxiv.org/abs/2602.22755},
  language   = {en}
}

@misc{kadavath_language_2022,
  title     = {Language models (mostly) know what they know},
  author    = {Kadavath, Saurav and Conerly, Tom and Askell, Amanda and Henighan, Tom and Drain, Dawn and Perez, Ethan and Schiefer, Nicholas and Hatfield-Dodds, Zac and DasSarma, Nova and Tran-Johnson, Eli and Johnston, Scott and El-Showk, Sheer and Jones, Andy and Elhage, Nelson and Hume, Tristan and Chen, Anna and Bai, Yuntao and Bowman, Sam and Fort, Stanislav and Ganguli, Deep and Hernandez, Danny and Jacobson, Josh and Kernion, Jackson and Kravec, Shauna and Lovitt, Liane and Ndousse, Kamal and Olsson, Catherine and Ringer, Sam and Amodei, Dario and Brown, Tom and Clark, Jack and Joseph, Nicholas and Mann, Ben and McCandlish, Sam and Olah, Chris and Kaplan, Jared},
  year      = 2022,
  month     = nov,
  publisher = {arXiv},
  doi       = {10.48550/arXiv.2207.05221},
  url       = {http://arxiv.org/abs/2207.05221},
  urldate   = {2026-05-16},
  language  = {en}
}

@misc{zou_representation_2025,
  title      = {Representation engineering: a top-down approach to {AI} transparency},
  shorttitle = {Representation engineering},
  author     = {Zou, Andy and Phan, Long and Chen, Sarah and Campbell, James and Guo, Phillip and Ren, Richard and Pan, Alexander and Yin, Xuwang and Mazeika, Mantas and Dombrowski, Ann-Kathrin and Goel, Shashwat and Li, Nathaniel and Byun, Michael J. and Wang, Zifan and Mallen, Alex and Basart, Steven and Koyejo, Sanmi and Song, Dawn and Fredrikson, Matt and Kolter, J. Zico and Hendrycks, Dan},
  year       = 2025,
  month      = mar,
  publisher  = {arXiv},
  doi        = {10.48550/arXiv.2310.01405},
  url        = {http://arxiv.org/abs/2310.01405},
  urldate    = {2026-05-16},
  copyright  = {arXiv.org perpetual, non-exclusive license},
  language   = {en}
}

@inproceedings{rimsky_steering_2024,
  title     = {Steering llama 2 via contrastive activation addition},
  author    = {Rimsky, Nina and Gabrieli, Nick and Schulz, Julian and Tong, Meg and Hubinger, Evan and Turner, Alexander},
  year      = 2024,
  booktitle = {Proceedings of the 62nd {Annual} {Meeting} of the {Association} for {Computational} {Linguistics} ({Volume} 1: {Long} {Papers})},
  publisher = {Association for Computational Linguistics},
  address   = {Bangkok, Thailand},
  pages     = {15504--15522},
  doi       = {10.18653/v1/2024.acl-long.828},
  url       = {https://aclanthology.org/2024.acl-long.828},
  urldate   = {2026-05-16},
  language  = {en}
}

@misc{dietz_split_2026,
  title      = {Split personality training: revealing latent knowledge through alternate personalities},
  shorttitle = {Split personality training},
  author     = {Dietz, Florian and Wale, William and Gilg, Oscar and McCarthy, Robert and Michalak, Felix and Danon, Gustavo Ewbank Rodrigues and Guzman, Miguelito de and Klakow, Dietrich},
  year       = 2026,
  doi        = {10.48550/arXiv.2602.05532},
  url        = {https://arxiv.org/abs/2602.05532},
  language   = {en}
}

@misc{ravindran_adversarial_2025,
  title      = {Adversarial activation patching: a framework for detecting and mitigating emergent deception in safety-aligned transformers},
  shorttitle = {Adversarial activation patching},
  author     = {Ravindran, Santhosh Kumar},
  year       = 2025,
  doi        = {10.48550/arXiv.2507.09406},
  url        = {https://arxiv.org/abs/2507.09406},
  language   = {en}
}

@misc{basu_interpretability_2026,
  title      = {Interpretability without actionability: mechanistic methods cannot correct language model errors despite near-perfect internal representations},
  shorttitle = {Interpretability without actionability},
  author     = {Basu, Sanjay and Patel, Sadiq Y. and Sheth, Parth and Muralidharan, Bhairavi and Elamaran, Namrata and Kinra, Aakriti and Morgan, John and Batniji, Rajaie},
  year       = 2026,
  doi        = {10.48550/arXiv.2603.18353},
  url        = {https://arxiv.org/abs/2603.18353},
  language   = {en}
}

@inproceedings{platt_probabilistic_1999,
  title     = {Probabilistic outputs for support vector machines and comparisons to regularized likelihood methods},
  author    = {Platt, John C.},
  year      = 1999,
  booktitle = {Advances in {Large} {Margin} {Classifiers}},
  publisher = {MIT Press},
  pages     = {61--74},
  language  = {en}
}

@inproceedings{zadrozny_transforming_2002,
  title     = {Transforming classifier scores into accurate multiclass probability estimates},
  author    = {Zadrozny, Bianca and Elkan, Charles},
  year      = 2002,
  booktitle = {Proceedings of the {Eighth} {ACM} {SIGKDD} {International} {Conference} on {Knowledge} {Discovery} and {Data} {Mining}},
  publisher = {ACM},
  pages     = {694--699},
  doi       = {10.1145/775047.775151},
  language  = {en}
}

@inproceedings{kull_beyond_2017,
  title      = {Beyond sigmoids: how to obtain well-calibrated probabilities from binary classifiers with beta calibration},
  shorttitle = {Beyond sigmoids},
  author     = {Kull, Meelis and Silva Filho, Telmo and Flach, Peter},
  year       = 2017,
  booktitle  = {Proceedings of the 20th {International} {Conference} on {Artificial} {Intelligence} and {Statistics}},
  publisher  = {PMLR},
  pages      = {623--631},
  doi        = {10.1214/17-ejs1338si},
  language   = {en}
}

@inproceedings{wang_self-consistency_2022,
  title     = {Self-consistency improves chain of thought reasoning in language models},
  author    = {Wang, Xuezhi and Wei, Jason and Schuurmans, Dale and Le, Quoc V. and Chi, Ed H. and Narang, Sharan and Chowdhery, Aakanksha and Zhou, Denny},
  year      = 2022,
  month     = sep,
  booktitle = {Proceedings of the 11th {International} {Conference} on {Learning} {Representations}},
  url       = {https://openreview.net/forum?id=1PL1NIMMrw},
  urldate   = {2026-05-18},
  language  = {en}
}

@article{lin_teaching_2022,
  title    = {Teaching models to express their uncertainty in words},
  author   = {Lin, Stephanie and Hilton, Jacob and Evans, Owain},
  year     = 2022,
  month    = jun,
  journal  = {Transactions on Machine Learning Research},
  pages    = {1--19},
  issn     = {2835-8856},
  url      = {https://openreview.net/forum?id=8s8K2UZGTZ},
  urldate  = {2026-05-18},
  language = {en}
}

@inproceedings{guo_calibration_2017,
  title     = {On calibration of modern neural networks},
  author    = {Guo, Chuan and Pleiss, Geoff and Sun, Yu and Weinberger, Kilian Q.},
  year      = 2017,
  month     = jul,
  booktitle = {Proceedings of the 34th {International} {Conference} on {Machine} {Learning}},
  publisher = {PMLR},
  pages     = {1321--1330},
  issn      = {2640-3498},
  url       = {https://proceedings.mlr.press/v70/guo17a.html},
  urldate   = {2026-05-18},
  note      = {shortConferenceName: ICML},
  language  = {en}
}

@inproceedings{kuhn_semantic_2022,
  title      = {Semantic uncertainty: linguistic invariances for uncertainty estimation in natural language generation},
  shorttitle = {Semantic uncertainty},
  author     = {Kuhn, Lorenz and Gal, Yarin and Farquhar, Sebastian},
  year       = 2022,
  month      = sep,
  booktitle  = {Proceedings of the 11th {International} {Conference} on {Learning} {Representations}},
  url        = {https://openreview.net/forum?id=VD-AYtP0dve},
  urldate    = {2026-05-18},
  language   = {en}
}

@misc{zur_are_2025,
  title      = {Are language models aware of the road not taken? {Token}-level uncertainty and hidden state dynamics},
  shorttitle = {Are language models aware of the road not taken?},
  author     = {Zur, Amir and Geiger, Atticus and Lubana, Ekdeep Singh and Bigelow, Eric},
  year       = 2025,
  month      = nov,
  publisher  = {arXiv},
  doi        = {10.48550/arXiv.2511.04527},
  url        = {http://arxiv.org/abs/2511.04527},
  urldate    = {2026-05-18},
  note       = {arXiv:2511.04527 [cs.CL]},
  language   = {en}
}

@inproceedings{jakkli_current_2026,
  title     = {Current Activation Oracles Are Hard to Use on Safety-Relevant Tasks},
  author    = {Jakkli, Arya and Rajamanoharan, Senthooran and Nanda, Neel},
  year      = 2026,
  booktitle = {Mechanistic Interpretability Workshop at the 43rd International Conference on Machine Learning},
  url       = {https://openreview.net/forum?id=7nRmqgz3Wv}
}

@inproceedings{bauer_building_2026,
  title     = {Building Better Activation Oracles},
  author    = {Bauer, Jan and De Schamphelaere, Celeste and Karvonen, Adam and Luick, Niclas and Nanda, Neel},
  year      = 2026,
  booktitle = {Mechanistic Interpretability Workshop at the 43rd International Conference on Machine Learning},
  url       = {https://arxiv.org/abs/2606.02609}
}

@misc{cywinski_eliciting_2025,
  title         = {Eliciting Secret Knowledge from Language Models},
  author        = {Cywi{\'n}ski, Bartosz and Ryd, Emil and Wang, Rowan and Rajamanoharan, Senthooran and Nanda, Neel and Conmy, Arthur and Marks, Samuel},
  year          = 2025,
  url           = {https://arxiv.org/abs/2510.01070},
  eprint        = {2510.01070},
  archiveprefix = {arXiv},
  primaryclass  = {cs.LG}
}

@misc{hu_lora_2021,
  title         = {{LoRA}: Low-Rank Adaptation of Large Language Models},
  author        = {Hu, Edward J. and Shen, Yelong and Wallis, Phillip and Allen-Zhu, Zeyuan and Li, Yuanzhi and Wang, Shean and Wang, Lu and Chen, Weizhu},
  year          = 2021,
  url           = {https://arxiv.org/abs/2106.09685},
  eprint        = {2106.09685},
  archiveprefix = {arXiv},
  primaryclass  = {cs.CL}
}

@article{mielke_reducing_2022,
  title    = {Reducing conversational agents’ overconfidence through linguistic calibration},
  volume   = {10},
  issn     = {2307-387X},
  url      = {https://doi.org/10.1162/tacl_a_00494},
  doi      = {10.1162/tacl_a_00494},
  language = {en},
  urldate  = {2026-07-15},
  journal  = {Transactions of the Association for Computational Linguistics},
  author   = {Mielke, Sabrina J. and Szlam, Arthur and Dinan, Emily and Boureau, Y-Lan},
  month    = aug,
  year     = {2022},
  pages    = {857--872}
}

@article{xiong_can_2024,
  title      = {Can {LLMs} express their uncertainty? {An} empirical evaluation of confidence elicitation in {LLMs}},
  volume     = {2024},
  shorttitle = {Can {LLMs} express their uncertainty?},
  url        = {https://proceedings.iclr.cc/paper_files/paper/2024/hash/6733cf15e10e2cd1d59af033c3bb8507-Abstract-Conference.html},
  language   = {en},
  urldate    = {2026-07-15},
  journal    = {International Conference on Learning Representations},
  author     = {Xiong, Miao and Hu, Zhiyuan and Lu, Xinyang and Li, Yifei and Fu, Jie and He, Junxian and Hooi, Bryan},
  month      = may,
  year       = {2024},
  pages      = {23650--23678}
}

@misc{nostalgebraist_logit_2020,
  title        = {interpreting {GPT}: the logit lens},
  author       = {{nostalgebraist}},
  year         = 2020,
  howpublished = {LessWrong},
  url          = {https://www.lesswrong.com/posts/AcKRB8wDpdaN6v6ru/interpreting-gpt-the-logit-lens}
}

@misc{pan_latentqa_2024,
  title         = {{LatentQA}: Teaching {LLMs} to Decode Activations Into Natural Language},
  author        = {Pan, Alexander and Chen, Lijie and Steinhardt, Jacob},
  year          = 2024,
  url           = {https://arxiv.org/abs/2412.08686},
  eprint        = {2412.08686},
  archiveprefix = {arXiv},
  primaryclass  = {cs.CL}
}

@inproceedings{minder_narrow_2026,
  title     = {Narrow Finetuning Leaves Clearly Readable Traces in Activation Differences},
  author    = {Minder, Julian and Dumas, Cl{\'e}ment and Slocum, Stewart and Casademunt, Helena and Holmes, Cameron and West, Robert and Nanda, Neel},
  year      = 2026,
  booktitle = {International Conference on Learning Representations},
  url       = {https://arxiv.org/abs/2510.13900}
}

@article{owen_pigeonhole_2007,
  title   = {The pigeonhole bootstrap},
  author  = {Owen, Art B.},
  journal = {The Annals of Applied Statistics},
  volume  = {1},
  number  = {2},
  year    = {2007},
  doi     = {10.1214/07-AOAS122}
}

@inproceedings{holtzman_surface_2021,
  title     = {Surface Form Competition: Why the Highest Probability Answer Isn't Always Right},
  author    = {Holtzman, Ari and West, Peter and Shwartz, Vered and Choi, Yejin and Zettlemoyer, Luke},
  booktitle = {Proceedings of the 2021 Conference on Empirical Methods in Natural Language Processing},
  pages     = {7038--7051},
  publisher = {Association for Computational Linguistics},
  year      = {2021},
  doi       = {10.18653/v1/2021.emnlp-main.564}
}

@inproceedings{geifman_selective_2017,
  title = {Selective Classification for Deep Neural Networks},
  author = {Geifman, Yonatan and El-Yaniv, Ran},
  booktitle = {Advances in Neural Information Processing Systems},
  volume = {30},
  year = {2017},
  url = {https://proceedings.neurips.cc/paper_files/paper/2017/hash/4a8423d5e91fda00bb7e46540e2b0cf1-Abstract.html},
}
